\documentclass[10pt,journal,compsoc]{IEEEtran}

\ifCLASSOPTIONcompsoc
  \usepackage[nocompress]{cite}
\else
  \usepackage{cite}
\fi

\ifCLASSINFOpdf
\else
\fi

\usepackage{booktabs}
\usepackage{graphicx}
\usepackage{cite}

\usepackage{bm}
\usepackage{amssymb}
\usepackage{latexsym,amscd}

\usepackage{algorithm}
\usepackage{algorithmic}
\usepackage{graphicx}
\usepackage{float}
\usepackage{stfloats}
\usepackage{epstopdf}
\usepackage{color}
\usepackage{multirow}
\usepackage{subfigure} 
\usepackage{amsfonts} 

\usepackage{booktabs}  
\usepackage{threeparttable}  

\usepackage{CJK}
\usepackage{amsmath}

\usepackage[T1]{fontenc}

\usepackage{url}

\begin{document}

\title{Concept Drift and Long-Tailed Distribution in Fine-Grained Visual Categorization: Benchmark and Method}

\author{Shuo~Ye, Shiming Chen, Ruxin Wang, Tianxu Wu, Salman Khan,~\IEEEmembership{Senior Member,~IEEE}, Fahad Shahbaz Khan, and Ling Shao, ~\IEEEmembership{Fellow,~IEEE}, 

\thanks{Corresponding author: \textit{Shiming Chen. e-mail: gchenshiming@gmail.com.} } 

\thanks{Shuo~Ye and Tianxu Wu are with the School of Electronic Information and Communications, Huazhong University of Science and Technology.}

\thanks{Ruxin Wang is with the Alibaba Group, DEEPROUTE.AI.}

\thanks{Shiming Chen is with Mohamed bin Zayed University of AI.}

\thanks{Salman Khan is with Mohamed bin Zayed University of AI and Australian National University.}

\thanks{Fahad Shahbaz Khan is with Mohamed bin Zayed University of AI and Linkoping University.}

\thanks{L. Shao is with UCAS-Terminus AI Lab, University of Chinese Academy of Sciences.}
}

\markboth{Journal of \LaTeX\ Class Files,~Vol.~14, No.~8, August~2015}%
{Shell \MakeLowercase{\textit{et al.}}: Bare Demo of IEEEtran.cls for Computer Society Journals}

\IEEEtitleabstractindextext{%

\begin{abstract}
Data is the foundation for the development of computer vision, and the establishment of datasets plays an important role in advancing the techniques of fine-grained visual categorization~(FGVC). In the existing FGVC datasets used in computer vision, it is generally assumed that each collected instance has fixed characteristics and the distribution of different categories is relatively balanced. In contrast, the real world scenario reveals the fact that the characteristics of instances tend to vary with time and exhibit a long-tailed distribution. Hence, the collected datasets may mislead the optimization of the fine-grained classifiers, resulting in unpleasant performance in real applications. Starting from the real-world conditions and to promote the practical progress of fine-grained visual categorization, we present a Concept Drift and Long-Tailed Distribution dataset. Specifically, the dataset is collected by gathering 11195 images of 250 instances in different species for 47 consecutive months in their natural contexts. The collection process involves dozens of crowd workers for photographing and domain experts for labeling. 
Meanwhile, we propose a feature recombination framework to address the learning challenges associated with CDLT. Experimental results validate the efficacy of our method while also highlighting the limitations of popular large vision-language models (e.g., CLIP) in the context of long-tailed distributions. This emphasizes the significance of CDLT as a benchmark for investigating these challenges.
The dataset is available at: \url{https://github.com/SYe-hub/CDLT}
\end{abstract}

}

\maketitle

\IEEEdisplaynontitleabstractindextext

%
\IEEEpeerreviewmaketitle

\section{Introduction}
\IEEEPARstart
Fine-grained visual categorization (FGVC) targets at accurately identifying the subordinate classes from the target class in which the instances exhibit highly similar appearances. This task is one of the most important components in computer vision systems and has been widely used in target recognition, human computer interaction, and robotics \cite{wei2021fine,wei2019rpc,merler2007recognizing,jund2016freiburg,liu2016deepfashion}. As an integral part of the research on computer vision, image datasets are the basis of data-driven models and play an important role in the development of FGVC. A large number of datasets are built up for algorithm design~(e.g., animals~\cite{parkhi2012cats}, plants~\cite{hou2017vegfru}, and vehicles~\cite{maji2013fine,krause20133d}).
While these datasets have witnessed the empirical improvements of the algorithm performance, the resultant algorithms and methods may still fail in real FGVC applications. One of the important reasons concerns about the misalignment between the distribution of the collected data and that of the whole data in real conditions (e.g., different time and different species). 

\begin{table*}[htbp]\small
\begin{center}
\caption{Summary of the popular FGVC datasets. "Source" refers to the acquisition manner of the data, i.e., collected from website or by photographing in real scenes. "Imbalance" represents the number of images in the largest class divided by the number of images in the smallest class. "Insts" refers to the number of all instances in the dataset. "BBox" and "Parts" refer to the object bounding box and the part annotations, respectively. "Texts" refers to whether fine-grained text descriptions are provided for each image. "n/a" denotes "not available".}
 \label{summarize}
 \begin{tabular}{cccccc}  
\toprule  \toprule   
  \textbf{Dataset}       &\textbf{Categories} &\textbf{Insts} & \textbf{Source} & \textbf{Annotations} & \textbf{Imbalance} \\  
\midrule   

Oxford flower17\cite{nilsback2006visual}        & 17   & 1360 	 & website          &	Texts 		& 1.00   \\
Oxford flower103\cite{nilsback2008automated}       & 103	 & 8189	 & website \& photo &	 Texts     & 6.25	\\
Stanford Dogs\cite{khosla2011novel}                    & 120  & 20580  & website          & BBox        & 1.00   \\ 
CUB 200-2011\cite{wah2011caltech}                    & 200  & 11788  & website          & Texts+BBox+Parts  & 1.03 \\ 
LeafSnap\cite{kumar2012leafsnap}                     & 184	 & 30866  & photo          & n/a         & 8.00	\\
Oxford Pets\cite{parkhi2012cats}                     & 37   & 7349	  & website         & BBox+Parts  & 1.08 \\   
Aircraft\cite{maji2013fine}			              & 100  & 10000  & photo             & BBox        & 1.03  \\
Stanford Cars\cite{krause20133d}	                & 196  & 16185  & website           & BBox        & 2.83  \\
Food101\cite{bossard2014food}	 			      & 101  & 101000 & website            & BBox         &	1 \\
NABirds\cite{van2015building}		                & 555  & 48562  & website           & BBox+Parts  & 15.00 \\
DeepFashion\cite{liu2016deepfashion}             & 1050 & 800000 & website	       & BBox+Parts   & -  \\
Census Cars\cite{gebru2017bfine}	                 & 2675 & 712430 & website	       & BBox         & 10.00 \\
VegFru\cite{hou2017vegfru}			           & 292  & 160731 & website	       & BBox         & 8.00 \\
Urban Trees\cite{wegner2016cataloging}	        & 171  & 14572  & website         & n/a          & 7.51	\\
MVTec D2S\cite{follmann2018mvtec}                  & 60   & 21000	  & photo	       & BBox         &  1.83 \\
RPC\cite{wei2019rpc}				                   &200	 & 83739	  & photo	       & BBox         & 3.10  \\ \bottomrule  \bottomrule  
\end{tabular}
\end{center}
\vspace{-0.4cm}
\end{table*}

To see the above issue, we summarize the statistics of the commonly used FGVC datasets in Table \ref{summarize}, which reveals two trends along the development of FGVC.
First, along with time, the collected images tend to cover either increasing instances or increasing categories. 
In this development, many efforts have been devoted to pursuing more and more images to enrich the descriptions of instances, encouraging the data-driven models to extract more discriminative features for the instances. This is particularly important in FGVC since its performance heavily relies on the discriminability and the stability of the instance appearance in different environments. However, all the mentioned datasets unfortunately ignore the fact that environment may change the appearances of instances periodically or change their morphological states permanently, which is considered as the issue of~\textit{concept drift}. This exists widely in real world, particularly in all cases of biological growth and biological adaption such as human face, animals, and plants which have varied appearances in different ages and different environments. Thus, the existing FGVC methods usuallly have degenerated performance since the shifted visual features do not follow the distribution characterized by the dataset during model training. Another factor of this failure is that the existing models do not possess the ability of adapting to the changes of the instances in the deployed environment.

Second, the collected images tend to follow a comprehensive distribution of the real instances. An important viewpoint conveys that the instances in real world often follow a long-tailed distribution as the instance frequency of each class is generally different. As a result, a collected dataset could have a large number of under-represented classes and a few classes with more than sufficient instances \cite{chu2020feature}. 
However, this issue has not been considered in the widely used datasets \cite{khosla2011novel,wah2011caltech,maji2013fine}, but been noted in the recently published datasets \cite{hou2017vegfru,wei2019rpc} that collect data from real world. Considering that the Internet images are often prettified in which case the included instances exhibit healthy, tidy, and gorgeous appearances,
which may restrict the diversity of the instance images. This brings the issue that the imbalance in the distribution of collected data could not adequately describe the challenge of the real world long-tailed distribution.
Please note that in the real world, the challenges posed by long-tailed distributions are even more complex than anticipated.
Especially with the rise of large vision-language models (VLMs)~\cite{liu2024survey}, large-scale web-based image-text pretraining datasets encompass a wide range of visual concepts and provide abundant prior knowledge for downstream tasks. VLMs such as CLIP play a critical role in contemporary deep learning.
However, downstream tasks often inherit biases present in the pretraining data~\cite{mehrabi2021survey}. The models tend to rely heavily on feature patterns from common concepts, limiting VLMs' ability to generalize to rare categories, particularly in datasets with long-tailed distributions~\cite{zhao2024evaluating}.

The above two trends and the associated issues prompt us to a direct question: how could we deploy the research result in real FGVC applications and what is the obstacle that prevents us to accomplish this job? Targeting at this, in this paper, we introduce a new dataset, termed as Concept Drift and Long-Tailed Distribution~(CDLT), which focuses on the two mentioned issues, i.e. concept drift and long-tailed distribution, and promotes the development of practical FGVC methods. Before collecting data, we find a species named as succulent plant, which strikes a good trade-off between the accessibility and the periodic variation speed. During collection, we realize the concept drift issue by setting the time span of collection as several consecutive years, where the data from different seasons are collected. Regarding the long-tailed distribution issue, we expand the geographic scale and the scene of collection to reflect the actual distribution of the data in different areas and different conditions.
Based on the collected data, the annotation about the season of each instance is also recorded, which facilitates a coherent description of the instance. 

The main contributions are summarized as follows:

\begin{itemize}
\item A novel FGVC dataset is proposed which, to the best of our knowledge, is the first fine-grained dataset concerning the concept drift and the long-tailed distribution issues in real world, which could facilitate the development of advanced FGVC algorithms.
\item A feature recombination framework is proposed to mitigate learning challenges arising from concept drift and long-tailed distributions in FGVC. Model performance is improved through a data augmentation strategy, particularly with effective enhancement of tail data. 
\item Experimental results demonstrate that our method establishes a new state-of-the-art performance on the CDLT, highlighting the limitations of popular large vision-language models (e.g., CLIP) on long-tailed distributions. This underscores the importance of our dataset as a benchmark for studying these challenges.
\end{itemize}
\vspace{-0.2cm}

The rest of this paper is organized as follows. Section 2 introduces the background knowledge. In Section 3, we discuss the details of CDLT. In Section 4, we propose a feature recombination method to alleviate learning biases caused by insufficient feature representations and imbalanced distributions in the feature space. The performance of several typical algorithms on our dataset and the performance of large vision-language models on long-tailed distributions will be presented in Section 5. Finally, Section 6 concludes this paper.
·

\section{Related Work}
In this section, we first present the formal definition of concept drift and introduce the common types in Section A. The preliminary knowledge of long-tailed effect in the real-world data is then reviewed in Section B.
\vspace{-0.1cm}
\subsection{Concept Drift}
Concept drift describes the unforeseeable changes in the underlying distribution of streaming data over time \cite{lu2018learning}, which can be formally defined as follows. Consider a set of instances, which is denoted as $ S_{0,t} = \left\{d_0, ... , d_t \right\} $, where $t$ is a time period $[0, t]$ and $ d_{i} = (X_i, y_i) $ is an instance composing of the feature vector $X_i$ and its corresponding label $y_i$. Then, we suppose that $S_{0,t}$ follows a certain distribution $F_{0,t}(X, y)$. Concept drift occurs at the timestamp $t+1$, if $F_{0,t}(X, y) \neq F_{t+1,\infty}(X, y)$ or equivalently, $ {\exists t: P_{t}(X, y)\neq P_{t+1}(X, y)}$ \cite{lu2018learning}.

\begin{figure}[htbp]
  \begin{center}
  \includegraphics[width=0.49\textwidth]{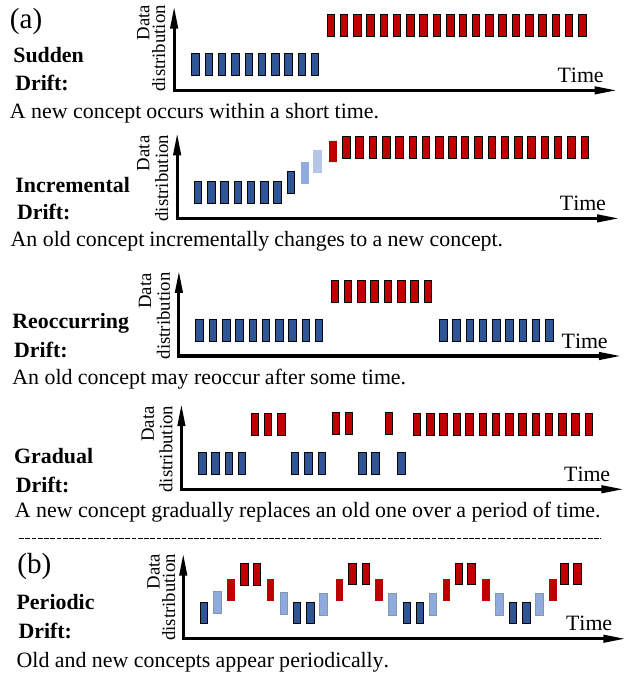}
  \end{center}
	\setlength{\abovecaptionskip}{-0.2cm} 
   \caption{Examples of different concept drift types. Red and blue represent two different concepts, while the light color represents the intermediate concept which appears during the transformation from one to another.}
\vspace{-0.2cm}
  \label{concept_drift}
\end{figure}

In a general sense, concept drift can be distinguished as four types \cite{gama2014survey} according to the variation form of the instance, including sudden, incremental, recurring, and gradual, as shown in Figure~\ref{concept_drift}(a). Sudden means that the appearance changes rapidly while the result is irreversible. Incremental indicates that the changes happen gradually and require a transition duration. Recurring denotes a temporary change that may be reversed to the former state of the instance within a short period of time which is, therefore, also referred as local drift. Note that recurring can be repeated in irregular time intervals without periodicity. Gradual means that in the process of replacing the old concepts with the new ones, the two concepts arise alternately. The current datasets focusing on concept drift can be divided into two categories according to the data source, i.e., synthetic datasets and real-world datasets. The synthetic datasets can simulate arbitrary types of drift which are, however, biased with respect to the real-world shifts. Specifically, the real-world data 1) do not exhibit the precise start and end time of the drifts, 2) usually contain mixed drift types, 3) contain temporal concept drift spanning over different durations such as daily, yearly \cite{elwell2011incremental}, and 4) target at a specific drift type such as Spam \cite{katakis2009adaptive}. 

In the data-driven frameworks, the simplest and most effective way to deal with concept drift is to fine-tune the classifier based on the drifted data. In details, a sliding window is designed to hold the latest data, which is used to optimize the model when the error rate increases \cite{bu2017incremental}. While concept drift occurs in stream data, the task of image recognition may face the same condition, that is, the photographed instance could vary the appearance or the morphological states as mentioned in the above context. However, the traditional methods are unqualified to address this issue in image data. On the other hand, the successful FGVC algorithms heavily rely on the discriminability and the stability of the instance appearance in different environments, which do not fulfil the real conditions. Therefore, to facilitate the deployment of FGVC in real-world, the data-driven model should endow prediction and decision-making with the adaptability in an uncertain environment. This poses the request of the dataset simulating different types of concept drift, promoting the research on this critical issue. To the best of our knowledge, we are the first to provide the concept drift dataset in FGVC, which contains a large variety of data expressing the changes of instances in 47 consecutive months. The types of drift not only include incremental and sudden, but also has periodic drift which is different from recurring, and instead indicates repeated conversion in regular intervals, as shown in Figure~\ref{concept_drift}(b).


\subsection{Long-tailed Effect}
There have already been several datasets with long-tailed distributions that are commonly used in computer vision, such as ImageNet-LT, Places-LT Dataset \cite{liu2019large}, LVIS \cite{gupta2019lvis}, Long-tailed CIFAR-10 and CIFAR-100 \cite{zhou2020bbn,cao2019learning}. Most of these datasets are created manually by sampling from the large datasets, for example, ImageNet-LT and Places-LT are collected by sampling a subset following the Pareto distribution from ImageNet \cite{deng2009imagenet} and Places \cite{zhou2017places}, respectively. 
In long-tailed distributions, the data imbalance between different classes and the biased coverage of the data distribution caused by limited data \cite{chu2020feature}. Especially for FGVC in which the sample numbers of the tail classes are inherently small. 
Note that for one category, all the samples of other categories including the background are regarded as negative. Hence, the rare categories can be easily overwhelmed by the majority categories during training and are inclined to be predicted as negatives \cite{tan2020equalization}. To alleviate this issue, the current researches mainly focus on the re-weighting strategy which assigns different weights to the losses of different instances to fully exploit the characteristics of the tail classes \cite{cui2019class,shu2019meta}. 

The above researches and datasets have promoted the solution of the long-tailed problem, but remain two issues that motivate us to build a new dataset in this work. Specifically, the first issue is that the existing methods over-focus on the tail classes, but ignore the similarity between the header and the tail classes. 
Addressing this issue is more challenging than anticipated. The classification performances of the previous methods are pleasing because the collected data describes hard boundaries \cite{yang2021delving} between classes where, for example, the class at the head is house and the class at the tail is tree. But in fine-grained identification, it would be fatal to ignore such similarity between the header and the tail data. Even the recently emerged VLMs struggle to effectively address this challenge. Despite being trained on large-scale web-based image-text datasets that encompass a wide range of visual concepts and provide abundant prior knowledge for downstream tasks, the biases inherent in this prior knowledge are inherited by downstream tasks and amplified in data with long-tailed distributions. We will demonstrate this in Section~(\ref{Ex_on_VLMs}).
The second issue states that the sampled datasets are difficult to simulate the real distribution in natural scenes, which hardly facilitates the deployment of FGVC algorithms in real applications. By contrast, the proposed dataset is collected from real-world scenarios, naturally exhibiting a real long-tailed distribution.


\begin{figure*}[htbp]
  \begin{center}
  \includegraphics[width=0.98\textwidth]{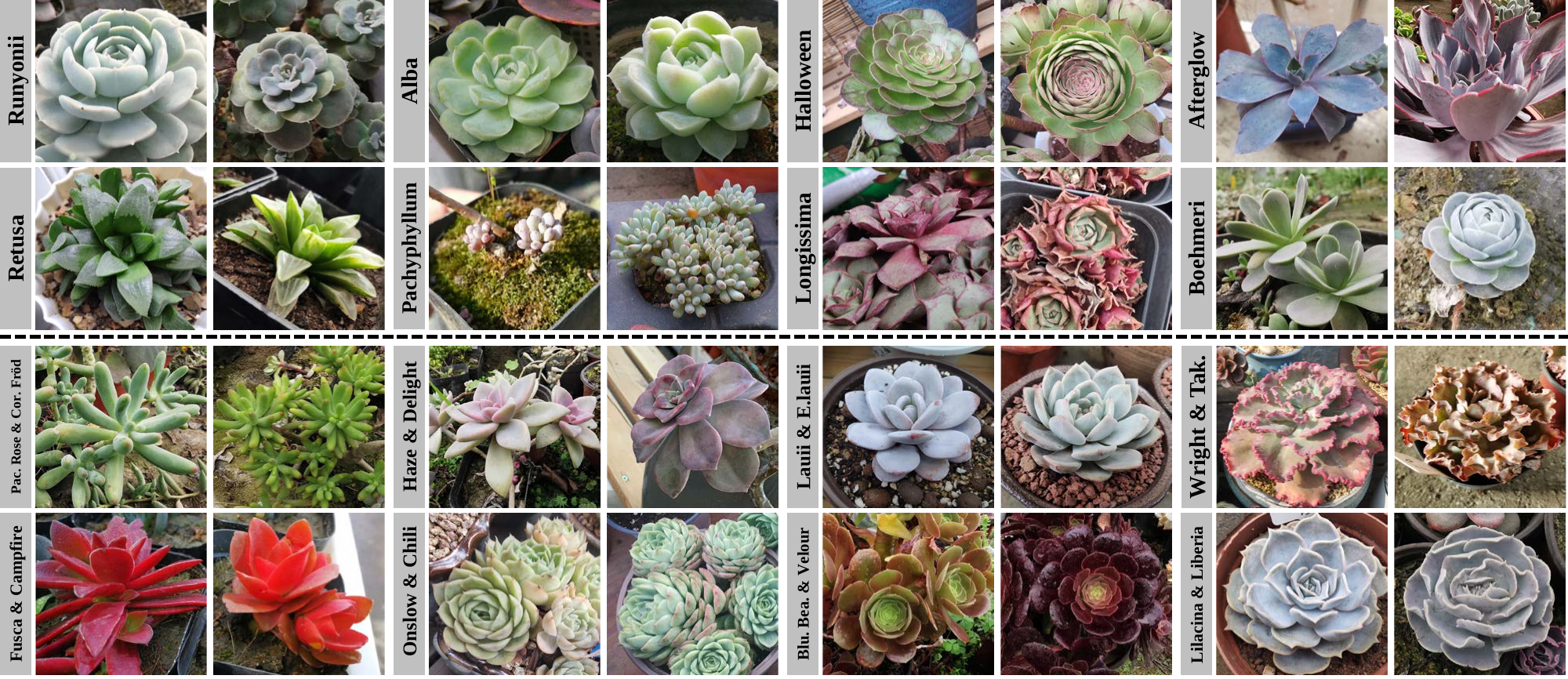}
  \end{center}
	\setlength{\abovecaptionskip}{-0.2cm} 
   \caption{Examples from the CDLT dataset. The first part shows the large variances in the same subcategory, and the second part illustrates the small variances among different subcategories.}
   \vspace{-0.2cm}
  \label{FGVC}
\end{figure*}

\section{Dataset Overview}
In this section, we describe the details of the CDLT dataset, including the instance characteristics, the collection process, and the splitting rule of the training and test data.

\subsection{Data Collection}

To build up a dataset that pursues the characteristics of real-world objects without any intentional crafting, we collect the data from 38 cities in 17 provincial-level administrative regions in China. The whole process lasts for 47 consecutive months, where the states of the objects in different seasons and different periods of a day are collected. This ensures that the data exhibits the concept drift phenomenon and the long-tailed effect, hence the dataset being named as CDLT, as shown in Figure~\ref{FGVC}. While the data comes from real world instead of websites, CDLT could promote the research on the state-of-the-art image classification for natural environment data. In the design of the dataset, each sample is expressed as the tuple of an image and its associated label. The collected raw images may contain outliers, hence remaining us a challenging task to make the dataset highly reliable. At this regard, we carefully process the candidate images through manual selection. In practice, the images of each sub-class are examined by four experts or longtime practitioners with identification experience. Only the images affirmed by more than two examiners are reserved, while the faded, blurry, and duplicated images are all filtered out. The label of an image consists of the descriptions on its family, genus, and species, which are derived mostly from Encyclopedia Illustration of Succulents Plants \cite{yan2019encyclopedia}. To enrich the label information, we also provide the auxiliary information which records the season of the instance at the photographing time. Note that the auxiliary information is different from multi-label or attribute \cite{patterson2012sun} as the former is not directly correlated with the semantic analysis of the instance identity. In total, we have completed the data construction of more than 6 families, 23 genus, and 258 subclasses, with more than 10,000 samples.

\begin{figure*}[htbp]
  \begin{center}
  \includegraphics[width=0.98\textwidth]{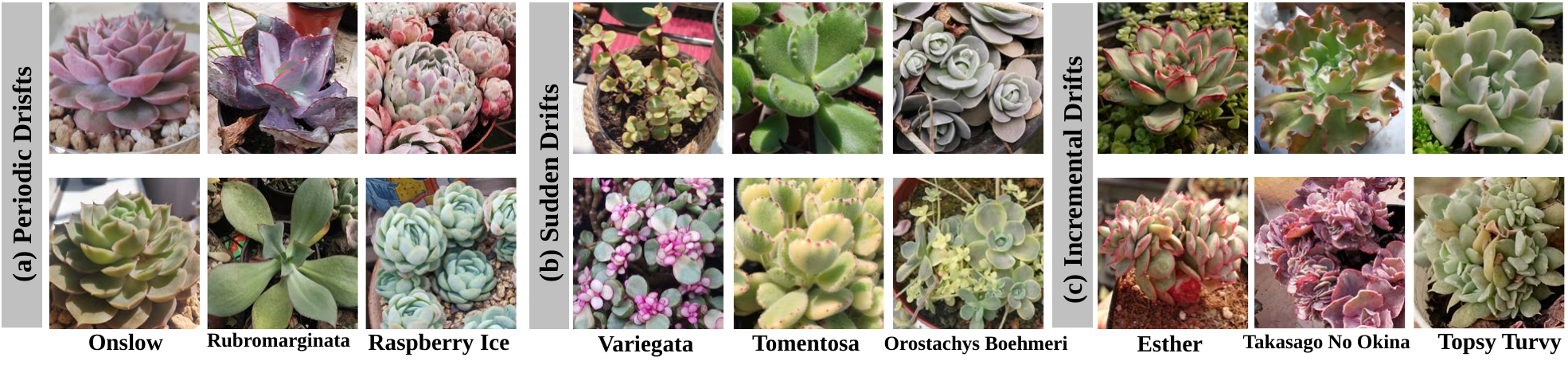}
  \end{center}
	\setlength{\abovecaptionskip}{-0.2cm} 
   \caption{Illustration of the concept drifts in CDLT.}
\vspace{-0.1cm}
  \label{concept_drift-all}
\end{figure*}
\subsection{Analyses on Concept Drift}
Consider that the interested objects in the current work belong to the same class of plants, which present large intra-class variance and small inter-class variance between subclasses. The difficulty of classification on these data is owing to the concept drift issues, including periodic, sudden, and incremental. 

Specifically, the periodic drift is represented by the variation of color, which means that the instances usually show periodic changes of color in different seasons, as shown in Figure~\ref{concept_drift-all}. While most of the classification methods (especially in FGVC) rely on the stability of the target appearance, color is one of the dominant characteristics that ensures the discriminability between different classes, e.g. birds vs. dogs. However, the periodic concept drift, which is widespread in families and genera, destroys this stability and poses a challenge to FGVC.
The sudden drift is also revealed by color which is, however, different from the periodic drift. On one hand, the variation in sudden drift is permanent, meaning that the changed state is not appeared in the history of the instance. On the other hand, one characteristic of this drift is partial, indicating that the color change only happen in parts of the instance appearance, while the rest remains the same as the original color.
The collected data relating to sudden drift is limited and unfortunately, the data augmentation techniques for enriching the samples of such a drift may be problematic. This is because by changing the ratio of colors in the color channel may lead to false textures, which poses unpredictable risks to recognition.
The incremental drift in CDLT is represented by the change of morphological characteristics.
For example, this drift can be demonstrated by the increasing number or the increasing size of leaves, when the leaves grow up along the time. In this process, the structural characteristics of the instance evolve to a different state, which could pose a great difficulty of recognition even for the human experts. The clues for discriminating such a kind of data could appear in the tip parts of the leaves, suggesting that certain localization methods are needed for accurate classification.

The natural environments generally drive the plants to be changed in complex conditions. This informs us that the collected data in CDLT exhibits a mixture of multiple concept drifts instead of only one drift. 
The concurrence of multiple concept drifts can bring more challenges to the design of advanced FGVC algorithms. 

\subsection{Analyses on Long-tailed Distribution}
\begin{figure}[htbp]
  \begin{center}
  \includegraphics[width=0.5\textwidth]{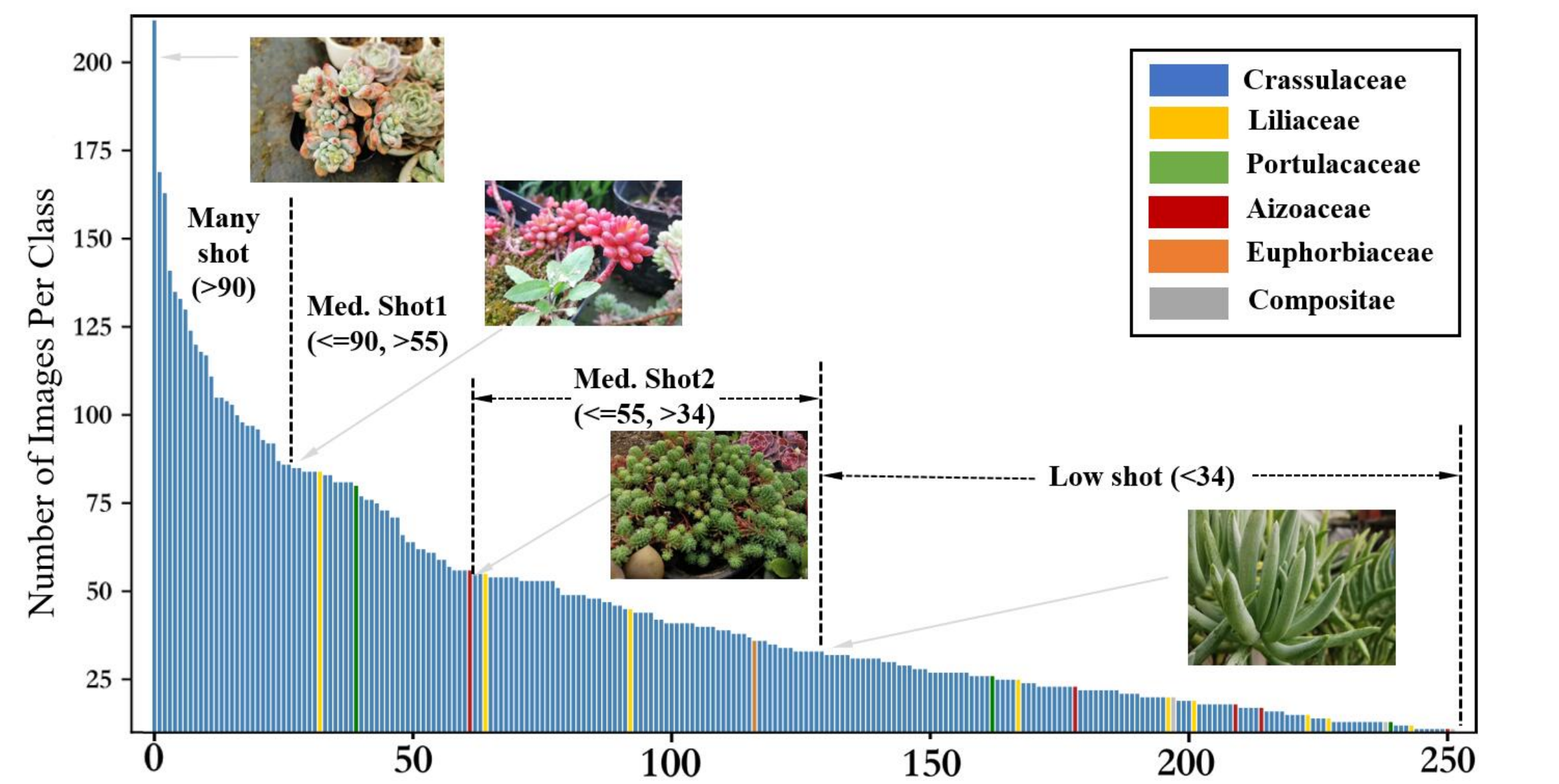}
  \end{center}
	\setlength{\abovecaptionskip}{-0.2cm} 
   \caption{The data distribution of common FGVC datasets.}
\vspace{-0.1cm}
  \label{distribution}
\end{figure}

The planting scale of the succulent plants follows the rule of the local market in China. In the general sense, the popular plants have a large planting scale while the others occupy a niche market. This affects the collection process, in which we acquire a large amount of, say, \textit{Lime\&Chili} and a small amount of e.g. \textit{Pachyphyllum Rose}. Therefore, the collected CDLT dataset naturally exhibits an imbalance distribution among the subclasses, as visualized in Figure~\ref{distribution}. Recall that each sample in CDLT is labelled according to families, genera, and species, where each species is a subclass corresponding to a column in Figure~\ref{distribution}, and different genera are distinguished by different colors. We rank the total number of samples in each subclass from the highest to the lowest, revealing a clear long-tailed distribution. It can be seen that CDLT has a great imbalance, which is not only manifested in species but also exists in genera. In addition, the frequency curve of the data is not very smooth, and is difficult to be approximated by a formalized distribution. An improper assumption on the distribution may lead to an unexpected impact on the final identification.

\subsection{Dataset Construction}

To construct a formal dataset, the collected raw images are carefully cropped such that the image has a proper view of the instance(s). When multiple instances appear in an image, we follow that the cropped image contains no more than 3 instances that belong to the same subclass. In this way, we obtain 14512 samples in total. Rather than releasing all the samples, 11195 samples are selected to be included in the final dataset, while each sub-class contains at least 10 images. The remaining samples are held in reserve in case we encountered unforeseen problems when using this dataset. All the images are resized to have a maximal dimension of 1200px.

\begin{figure}[htbp]
  \begin{center}
  \includegraphics[width=0.49\textwidth]{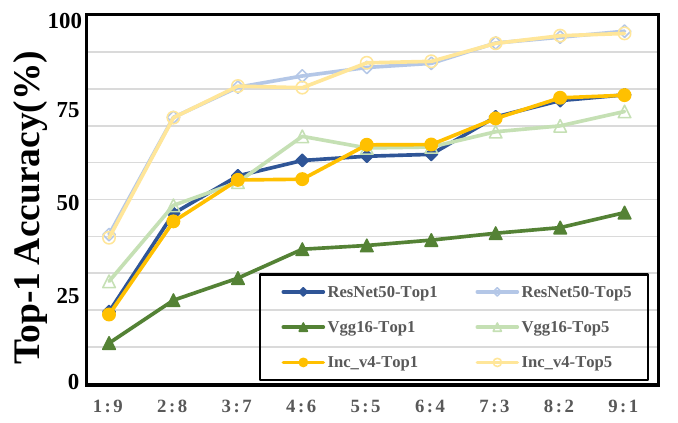}
  \end{center}
	\setlength{\abovecaptionskip}{-0.2cm} 
   \caption{Performance of the deep models under different partitioning ratios on CDLT.}
\vspace{-0.1cm}
  \label{division}
\end{figure}

For a fair comparison on this dataset in future research, we provide a train/test split with an appropriate ratio which is empirically determined based on the performances of popular deep models including ResNet-50 \cite{he2016deep} and Inception-v4 \cite{szegedy2016inception}, and VGG-16 \cite{simonyan2014very}. 
The results are reported in Figure~\ref{division}, in which the abscissa represents the proportion of data between the training set and the test set and the ordinate represents the classification accuracy. To fully evaluate the performances of the models, the top-1/top-5 accuracies are presented. It is seen that ResNet-50 is the best performer on our dataset, while VGG-16 is far behind ResNet-50 and Inception-v4. For all the models, the performance is enhanced with the increase of the training data, and the growth rate keeps stable around 6:4. This informs us that the bias and the variance of the data may reach a good balance at this point, where the learning ability and the generalization ability can be both challenging for model design. Therefore, the proportion 6:4 is selected for dividing the CDLT dataset, which is also consistent with most of the other fine-grained datasets. Finally, a slight adjustment is applied to ensure that the training and test sets are representative of the variability on object numbers and background. The best accuracy with this proportion is around 87\%, which suggests that the data in CDLT is difficult to be distinguished and hence, effective models and training strategies are desired.
To validate the distinguishability of the data in CDLT, we employ the best performing model, ResNet-50, to conduct a random label experiment. Specifically, the label of each training sample is randomly reassigned to a non-ground-truth label, while the labels of the test data remain unchanged. This operation is to validate whether the model trained on such data could reach a bottom of performance and whether the label information in the provided CDLT is meaningful. As expected, the results show that the top-1/top-5 accuracies are 0.53\% and 2.18\%, respectively.

The dataset is further processed to clearly express the issues of based on this partition ratio, the training sets for concept drift and long-tailed distribution under the 6:4-split setting are given separately. 
This dataset offers two partitioning modes. The details are summarized in Table~\ref{specific_information}.

\begin{table}[htbp]\normalsize
  \centering  
  \begin{threeparttable}  
  \renewcommand\tabcolsep{12.0pt}  
   \caption{Dataset Overview. \textit{C}$_{train}$, \textit{C}$_{test}$ represent the number of training and testing images in the dataset respectively.} 
  \label{specific_information}  
    \begin{tabular}{c | c | cc}  
    \toprule  \toprule
	\textbf{}   & \textbf{Categories}   &\textbf{\textit{C}$_{train}$} & \textbf{\textit{C}$_{test}$}\\  
    \hline
CDLT      & 258  & 6737 & 4458\cr 
CDLT-cd   & 258  & 5091 & 4458\cr
     \toprule   \toprule  
    \end{tabular}  
    \end{threeparttable}  
\vspace{-0.2cm}  
\end{table} 

CDLT can be used for general FGVC tasks, while CDLT-cd is specifically designed for concept drift research. It includes two training partitions and one test partition: one training partition contains data from a single season, while the other covers data from the entire year. The test partition covers data from the entire year. 
Note that, for example, not every subclass has a significant periodic drift with the seasons. So the other subclasses in this training set, we only include divide out instances those with significant concept drift. 
To avoid discrepancies in results due to differences in the amount of training data, the data scale for both training modes in CDLT-cd needs to be consistent.
A basic usage approach for CDLT-cd is to use files with the "S1" suffix for training. In the comparative experiment, samples are first drawn from files with the "S1S2" suffix to match the scale of "S1," and then used for training.

\begin{figure}[htbp]
  \begin{center}
  \includegraphics[width=0.49\textwidth]{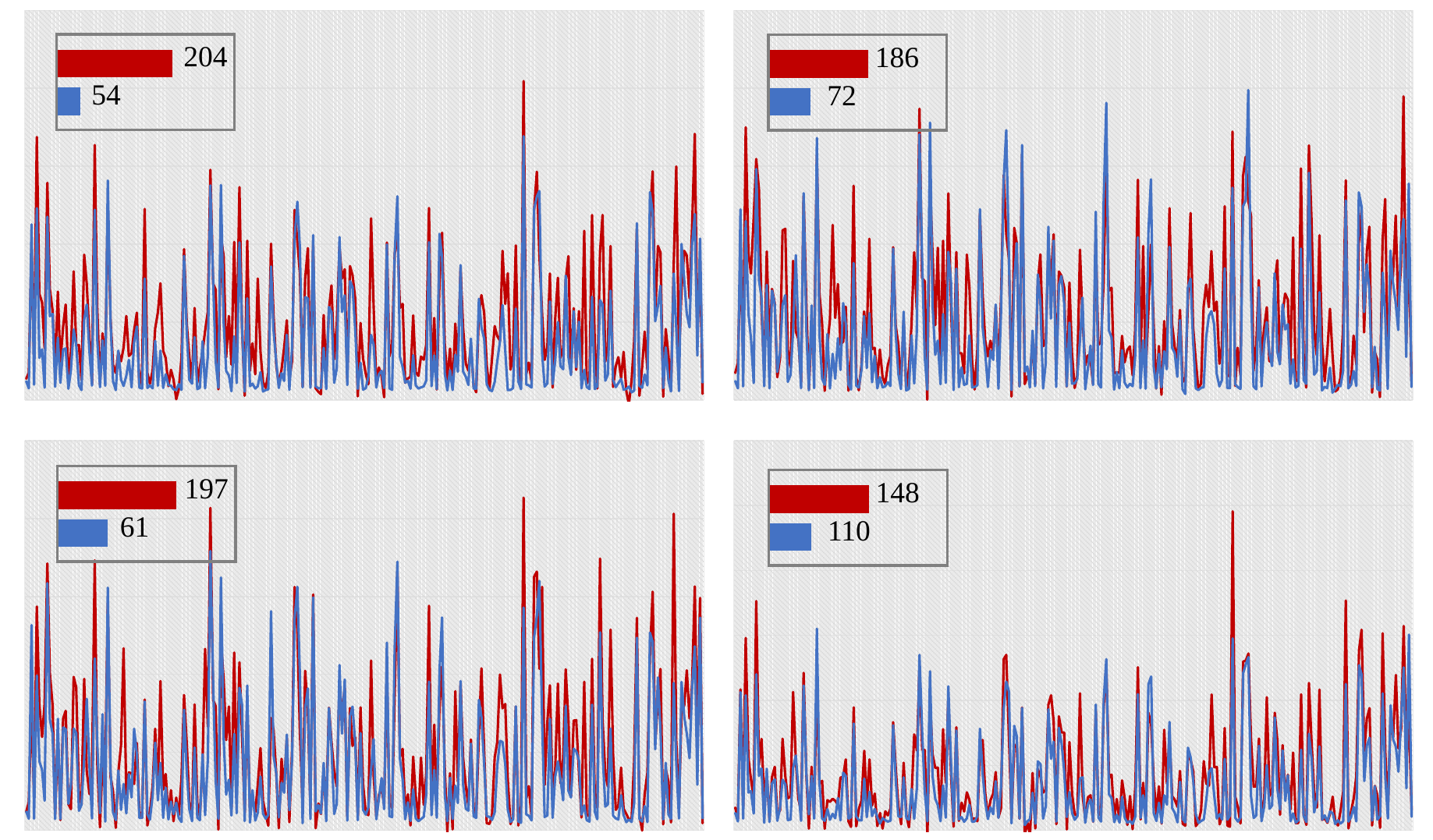}
  \end{center}
	\setlength{\abovecaptionskip}{-0.2cm} 
   \caption{Quantitative analysis of feature distribution under the ResNet-50 and ViT backbone networks. The red line represents the MMD between the training and test set, and the blue line represents the confidence values that accept the hypothesis.}
\vspace{-0.1cm}
  \label{CDLT_cd-MMD}
\end{figure}

To further justify the phenomena of concept drift in FGVC, we employ maximum mean discrepancy (MMD)\cite{gretton2012kernel} to quantitatively assess the concept drift in CDLT and CDLT-cd datasets.
The results are shown in Figure\ref{CDLT_cd-MMD}. 
The first and second rows show the results for ResNet-50 and ViT, respectively. The first column presents the results when training with data from a single season, while the second column shows the results when training with data from all seasons.
We compute the MMD values for each subclass between training and test sets, as shown in the red curve. The upper 95\% confidence interval bound (evaluated with 100 independent permutations) is shown as the blue line. We count the number of subclasses that the MMD value is above the upper 95\% confidence interval, which can be interpreted as having a "significant" distributional change. 
It can be found that concept drift is a universal problem in our dataset, as evidenced by different backbone networks. Training on data from a single season significantly impairs the model's ability to comprehensively understand the feature distribution, highlighting the challenges posed by the CDLT-cd dataset.
For the long long-tailed distribution issue problem, all instances samples of the training set can be usedare involved. In addition, we provide the season information of the instances samples are provided in the form of auxiliary information, which can be selectively used by users to improve the algorithm performance of the model according to their own needs. All instances are resized to have a max dimension of 1200px. The CDLT dataset is available from our project website. 

\section{Feature Recombination for FGVC}
To address the issues of concept drift and long-tailed distribution in FGVC, we propose a feature recombination-based FGVC framework. This framework aims to alleviate learning biases caused by insufficient feature representations and imbalanced distributions in the original feature space, enhancing the robustness and generalization of FGVC deep learning models.
\begin{figure*}[htbp]
  \begin{center}
  \includegraphics[width=0.98\textwidth]{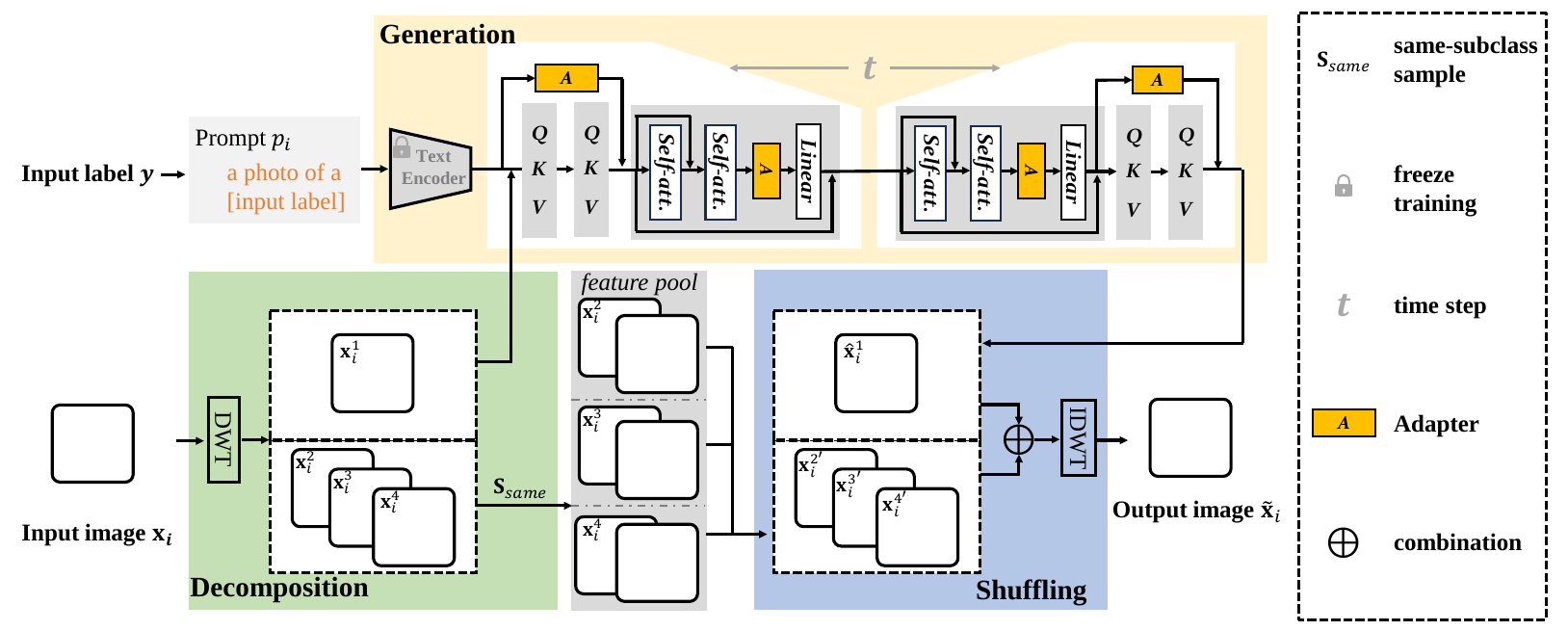}
  \end{center}
	\setlength{\abovecaptionskip}{-0.2cm} 
   \caption{Overview of the feature recombination framework, consisting of frequency decomposition, low-frequency image generation, high-frequency distribution shuffling and recombination.}
   \vspace{-0.2cm}
  \label{framework}
\end{figure*}
As shown in Figure\ref{framework}, this framework first decomposes the image into high/low-frequency parts using frequency transformation. For the low-frequency part, a diffusion generation strategy is employed to enhance the structural diversity of subclasses. For the high-frequency part, distribution shuffling is implemented to simulate the phenomenon of concept drift caused by changes in feature distributions. Subsequently, through feature recombination techniques, the feature representation in the original feature space is expanded, while addressing the challenge of FGVC caused by feature imbalanced distributions.


\subsection{Structural Augmentation}
Formally, $\textbf{x}_{i} \in \mathbb{R}^{h\times w\times c}$ is an image from the training dataset, $\mathcal{D}_g(\cdot) \colon \mathbb{R}^{h\times w\times c} \rightarrow \mathbb{R}^{h\times w\times c}$ denotes our method.
To obtain the final recombinated image $A_{i}$, input image $\textbf{x}_{i}$ goes through proposed generation using prompt $p_j$, distribution shuffling set $\mathcal{T}$. The overall augmentation proess can be represented as $A_{i} = \mathcal{D}_g (\textbf{x}_{i}, p_i, \mathcal{T})$\cite{zhou2022learning}.

{\bf Decomposition}: 
Frequency domain theory suggests that the global structural information of a spatial domain image is contained in the low-frequency band, while local textures are contained in the high-frequency band. Humans can easily recognize low-frequency images, whereas CNNs tend to fit high-frequency visual representations. However, when high-frequency components change, CNNs being overly sensitive to variations in high-frequency information, are prone to overfitting, leading to incorrect predictions~\cite{liu2024frequency}. 
Inspired by this, we employ DWT with the Haar wavelet flter by default to decompose the original input image $\textbf{x}_i$ into different frequency bands, and get the low-frequency subbands $\textbf{x}_{i}^{1}$ with the three high-frequency subbands $\textbf{x}_{i}^{2}$, $\textbf{x}_{i}^{3}$ and $\textbf{x}_{i}^{4}$.
Specifically, for a image $\textbf{x}$, row-wise and column-wise 1D-DWT are conducted, which are defined as 2D transform. The corresponding 2D flters are denoted as ${\textit{f}_{LL}, \textit{f}_{LH}, \textit{f}_{HL}, \textit{f}_{HH}}$ and the 2D flters of $\textit{f}_{LL}$ is formulated as:
\begin{equation}
\textit{f}_{LL} = \frac{1}{2} 
\begin{bmatrix}
1 & 1 \\
1 & 1
\end{bmatrix}.
\end{equation}
The $(i, j)$-th value of $\textbf{x}_{i}$ after the 2D Haar
transform can be calculated as:
\begin{equation}
\begin{aligned}
\textbf{x}_{i}^{1}(i, j) &= \frac{1}{2} \textbf{x}_{i}(2i - 1, 2j - 1) + \frac{1}{2} \textbf{x}_{i}(2i - 1, 2j) \\
&\quad + \frac{1}{2} \textbf{x}_{i}(2i, 2j - 1) + \frac{1}{2} \textbf{x}_{i}(2i, 2j).
\end{aligned}
\end{equation}
The remaining three high-frequency subbands $\textbf{x}_{i}^{2}$, $\textbf{x}_{i}^{3}$ and $\textbf{x}_{i}^{4}$ can also be defined in an analogous manner.
Our goal is to enhance low-frequency structural features while preserving high-frequency features as much as possible, in order to maintain the accuracy of the classification model.

{\bf Generation}: To obtain more structural features, low-frequency images $\textbf{x}_{i}^{1}$ of the same subclass are encoded and used to constrain the feature generation process of the LDM model, ensuring that the augmented low-frequency feature distribution by LDM remains consistent with the current subclass.
Specifically, in generation step $\mathcal{G}(\cdot)$, LDM takes the label of this subclass as a prompt $p_i$, expressed as "a photo of a [subclass label]", and The entire statement is sent to the pre-trained CLIP~\cite{radford2021learning} to transfer the text feature understanding. 

LDM is a type of latent variable models that include forward and reverse noise-injection process. During the forward process, noise is gradually added to the data, each step in the forward process is a Gaussian transition according to the following Markovian process:
\begin{equation}
q\left( \boldsymbol{x}_t|\boldsymbol{x}_{t-1} \right) =\mathcal{N} \left( \sqrt{\alpha _t}\boldsymbol{x}_{t-1},\beta _t\mathbf{I} \right) ,\forall t\in \left\{ 1,...,T \right\},
\end{equation}
\begin{equation}
q\left( \boldsymbol{x}_{1:T}|\boldsymbol{x}_{0} \right) =\prod_{t=0}^T{q\left( \boldsymbol{x}_t|\boldsymbol{x}_{t-1} \right)},
\end{equation}
where $T$ is the number of diffusion steps, $\boldsymbol{x}_t$ is the noise added version of the input $\textbf{x}_{i}^{1}$ with the timestamp $t$ and $\boldsymbol{x}_0=\textbf{x}_{i}^{1}$. The mean and variance of Gaussian noise are determined by $\beta_t$, $\alpha_{t}=1- \beta_t $. The reverse process is another Gaussian transition
\begin{equation}
    p_{\theta}\left( \boldsymbol{x}_{t-1}|\boldsymbol{x}_t \right) =\mathcal{N} \left( \boldsymbol{x}_{t-1}|\boldsymbol{\mu }_{\theta}\left( \boldsymbol{x}_t,t \right) ,\sigma _{\theta}^{2}\left( \boldsymbol{x}_t,t \right) \mathbf{I} \right),
\end{equation}
where the mean value $\boldsymbol{\mu }_{\theta}\left( \boldsymbol{x}_t,t \right) $ can be seen as the combination of  $\boldsymbol{x}_t$ and a noise prediction network $\epsilon _{\theta}\left( \boldsymbol{x}_t,t \right)$. The maximal likelihood estimation of the optimal mean is 
\begin{equation}
\tilde{\boldsymbol{\mu}}_{\theta}\left( \boldsymbol{x}_t,t \right) =\frac{1}{\sqrt{\alpha _t}}\left( \boldsymbol{x}_t-\frac{\beta _t}{\sqrt{1-\bar{\alpha}}}\mathbb{E} \left( \epsilon |\boldsymbol{x}_t \right) \right).
\end{equation}
To get the optimal mean, $\epsilon_{\theta}\left( \boldsymbol{x}_t,t \right)$ can be learned by a noise prediction objective
\begin{equation}
\min_{\theta} \mathbb{E} _{\boldsymbol{x}_0\sim q\left( \boldsymbol{x} \right) ,\epsilon \sim \mathcal{N} \left( 0,\mathbf{I} \right) ,t}\left\| \epsilon _{\theta}\left( \boldsymbol{x}_t,t \right) -\epsilon \right\| _{2}^{2}.
\end{equation}
However, in the context of conditional generation, the condition information $c$ should be considered during the training process of the noise prediction network
\begin{equation}
\min_{\theta} \mathbb{E} _{\boldsymbol{x}_0\sim q\left( \boldsymbol{x} \right) ,\epsilon \sim \mathcal{N} \left( 0,\mathbf{I} \right) ,t,c}\left\| \epsilon _{\theta}\left( \boldsymbol{x}_t,t,c \right) -\epsilon \right\| _{2}^{2}.
\end{equation}
In our method, $c$ is feature after CLIP.
The reconstruction loss of LDM is defined as follows:
\begin{equation}
\mathcal{L}_{LDM}=\mathbb{E}_{t,\boldsymbol {x}_0,c,\epsilon}||\epsilon_\theta\left(\boldsymbol{x}_t,t,c\right)-\epsilon||^2_{2}.
\end{equation}

Therefore, we got the augmented counterpart image $ \hat{\textbf {x}}_{i}^{1}$. 
The overall generation step can be represented as: $ \hat{ \textbf{x}}_{i}^{1} = \mathcal{G} (\textbf{x}_{i}^{1}, p_i)$.

\subsection{Distribution Shuffling}
{\bf Shuffling}: 
Since CNNs are easily misled by varying high-frequency information, we introduce a high-frequency distribution shuffling operation to maintain the model's basic classification performance while generating differentiated high-frequency features. 
Specifically, for a given training sample $(\mathbf{x_i}, y_i)$, it has three high-frequency subbands $\textbf{x}_{i}^{2}$, $\textbf{x}_{i}^{3}$ and $\textbf{x}_{i}^{4}$.

We record these three high-frequency subbands separately and establish a high-frequency feature set $\mathcal{T}$ for the current subclass. After obtaining the new low-frequency image data $\hat{ \textbf{x}}_{i}^{1}$, we randomly select $\textbf{x}_{i}^{2'}$, $\textbf{x}_{i}^{3'}$ and $\textbf{x}_{i}^{4'}$ from $\mathcal{T}$ to replace the high-frequency subband of $\textbf{x}_{i}$.
Finally, IDWT is employed to reconstructe each subbands from the wavelet domain to the spatial domain, denoted as:
\begin{equation}
 	\tilde{\textbf{x}}_{i} = \textrm {IDWT} (\hat{ \textbf{x}}_{i}^{1}, \textbf{x}_{i}^{2'}, \textbf{x}_{i}^{3'}, \textbf{x}_{i}^{4'}),
\end{equation}

\section{Experiments}
In this section, we empirically examine the characteristics of the proposed CDLT dataset by employing the SOTA FGVC methods. Quantitative analyses on the issues of long-tailed distribution and the concept drift are provided, with a final discussion on possible solutions in the future research on FGVC.

\subsection{Experimental Setting}
The real-world scenario of fine-grained classification is rather difficult compared with the existing fine-grained datasets for research, leading to the failure of the published FGVC algorithms in certain real-world cases. The proposed CDLT pursues such a difficulty, which is examined by performing several state-of-the-art deep models. Specifically, two commonly used backbones, including ResNets~\cite{he2016deep} and ViT~\cite{dosovitskiy2020image},
are tested on CDLT. Then, we also carry out a series of experiments on representative FGVC models, including B-CNN~\cite{zhu2017b}, HBP~\cite{yu2018hierarchical}, NTS-Net~\cite{Yang2018Learning}, MCL~\cite{chang2020devil}, and MOMN~\cite{min2020multi}. 
When employing the above-mentioned fine-grained models, we use two backbone networks to avoid the performance influence caused by the basic feature extractors, which are ResNet-50 and ViT pre-trained on ImageNet~\cite{russakovsky2015imagenet}. 

To investigate the long-tailed distribution problem, we further compare several SOTA methods that are specified for processing long-tailed data.
Including re-sampling method \cite{chawla2002smote,drumnond2003class}, focal loss~\cite{lin2017focal}, range loss~\cite{zhang2017range} and class-balanced loss~\cite{cui2019class}.
(1) The re-sampling method \cite{chawla2002smote,drumnond2003class} is typical to alleviate the long-tailed problem, which over-sample data from the minority categories \cite{chawla2002smote} and under-sample data from the majority categories \cite{drumnond2003class} such that the resampled data is balanced. (2) The focal loss \cite{lin2017focal} encourages hard negative mining, in which the weights of simple negative samples are reduced during training and the hard samples could be identified. (3) The range loss \cite{zhang2017range} aims at optimizing the harmonic mean values of the $k$ greatest range in one class. (4) The class-balanced loss \cite{cui2019class} considers the influence of the newly added data points and employs a re-weighting scheme to use the effective number of samples in each class for re-balance. 
All these methods are equipped with the same backbone networks (i.e., ResNet-50 and ViT) for a fair comparison. The general data enhancement techniques are used for effective learning. 
Finally, we evaluate the performance of each method to highlight their differences.
In the implementation of the re-sampling method, 20\% classes in the head and 20\% classes in the tail of the training set are used to calculate the mean values of the majority and the minority classes, respectively. Specifically, in CDLT, 80 and 13 are computed as the boundaries of the sample number in under-sampling and over-sampling, respectively. For the focal loss, we use different parameter settings for different backbones for preferred performance, that is, in ResNet-50 we let $\alpha$ = 0.5 and $\gamma = 2.0$. 
The range loss takes that $\alpha = 0.5$, $\gamma = 2.0$, $K = 2.0$, and $m = 4.0$. Regarding the class-balanced loss, 
The setting of the hyperparameters follows the published paper.

To avoid occasional factors in comparison, we repeat each experiment for three times independently and report the averaged performance. All the models take as input the images with the size of $600\times600$ pixels. The pre-processing of the input follows the common configurations. Specifically, random translation and random horizontal flipping are adopted for data augmentation during training and the center cropped image is used during inference. During training, we directly optimize the whole network using stochastic gradient descent with batch size of 16, momentum of 0.9, weight decay of $10^{-4}$ and learning rate of $10^{-3}$, periodically annealed by 0.1. The implementation is based on the Torch framework \cite{collobert2011torch7} and a Titan X GPU. 


\subsection{The Challenge of CDLT}
We first compare the basic deep models and the selected FGVC methods on CDLT to demonstrate the challenge of the proposed dataset. The results are listed in Table \ref{Results} in which top-1 accuracy is computed for each model. To make a reference of the performance, we evaluate all the selected models on CUB~\cite{wah2011caltech}, which is widely used in FGVC. 
As seen from the table, 
among the FGVC methods, B-CNN is a bad performer possibly because of the simple utilization of the discriminative features. HBP enhances the representation capability of bilinear pooling by incorporating multiple cross-layer bilinear features, thus achieving improved performance. The performance limitation of MOMN is due to the involvement of a complex optimization method which jointly optimizes three non-smooth constraints with different convex properties. The constraints may hinder the fitting procedure of MOMN on our dataset. The NTS-Net with ResNet-50, which considers the interview discriminant information, is the best performer among all the competitors. 
In contrast, our method achieves the best results, which can be attributed to the effective data augmentation that enables the model to learn the target features more comprehensively.
The orders of the performances of all the competitors on CDLT are mostly consistent with that on the CUB dataset, which validates the data effectiveness of CDLT. Moreover, the performance of the model on CDLT is lower than the corresponding case on CUB, demonstrating that CDLT brings a challenging classification task.

\begin{table}[htbp]\normalsize
  \centering  
  \begin{threeparttable}  
  \renewcommand\tabcolsep{12.0pt}  
   \caption{The classification performance of the SOTA FGVC methods on CDLT and CUB-200-2011.} 
  \label{Results}  
    \begin{tabular}{c | c | cc}  
    \toprule  \toprule
	\textbf{Method}  & \textbf{Backbone} & \textbf{CDLT}   &\textbf{CUB} \\  
    \hline
B-CNN\cite{zhu2017b}            & ResNet-50 & 72.79 & 83.88 \cr  
HBP\cite{yu2018hierarchical}    & ResNet-50 & 75.64 & 86.52 \cr
NTS-Net\cite{Yang2018Learning}  & ResNet-50 & 78.20 & 84.78 \cr  
MCL\cite{chang2020devil}        & ResNet-50 & 75.42 & 86.02 \cr   
MOMN\cite{min2020multi}         & ResNet-50 & 69.76 & 86.42 \cr
SPS~\cite{huang2021stochastic}  & ResNet-50 &-      & 87.29 \cr
OURS                            & ResNet-50 &\bf78.82&\bf87.85 \cr
    \hline
ViT~\cite{dosovitskiy2020image} & ViT-B-16  & 77.81  & 90.42 \cr 
FFVT~\cite{wang2021feature}     & ViT-B-16  &-       &91.62  \cr
OURS                            & ViT-B-16  &\bf79.20&\bf92.07 \cr
     \toprule   \toprule  
    \end{tabular}  
    \end{threeparttable}  
\vspace{-0.2cm}  
\end{table}

\subsection{Investigation on Long-tailed Distribution}
We investigate the impact of the long-tailed distribution issue in CDLT. Existing experiences tell that the classes with a majority of samples can cripple the feature extracting ability of a model on the tail classes, severely biasing the feature learning process \cite{ouyang2016factors}. To validate this fact in CDLT, we follow the classic manner that permutes the subclasses from the largest number of samples to the least number of samples. All the subclasses are grouped into four disjoint subsets according to the numbers of the class samples: many-shot classes (i.e. the classes each with over 90 training samples), medium-shot1 classes (i.e. the classes each with 55-90 training samples), medium-shot2 classes (i.e. the classes each with 34-55 training samples), and few-shot classes (i.e. the classes each with under 34 training samples). We divide the test set based on this grouping manner, in which way each subset accounts for 25\% of the test data. In each subset, the top-1 accuracies are reported in Table \ref{Long_tail}, which help us understand the detailed characteristics of each method.

\begin{table*}[htbp]
    \centering\normalsize
\begin{threeparttable}  
  \renewcommand\tabcolsep{3.5pt}  
\caption{The performance of the SOTA methods on long-tailed distributed data.}  
  \label{Long_tail} 
    \begin{tabular}{c|ccccc}
    \toprule  \toprule
                    & \multicolumn{5}{c}{\textbf{CDLT setting (ResNet-50)}} \\
               \textbf{Methods}& $>90$     & $\leq 90 \& >55$ & $\leq 55 \& >34$ & $<34$    \\
                    &\textbf{Many-shot} & \textbf{Medium-shot1} & \textbf{Medium-shot2} & \textbf{Few-shot} &\textbf{Overall} \\
    \hline
Softmax Loss\cite{hinton2015distilling}	& 73.97 & 72.27 & 53.24 & 48.89 & 62.27  \cr  
Over-sampling\cite{chawla2002smote}	    & 68.27 & 71.35 & 55.74 & 53.62 & 62.20  \cr  
Under-sampling                          & 41.10 & 49.31 & 43.53 & 50.46 & 46.21  \cr
Focal Loss\cite{lin2017focal}           & 75.90 & 75.94 & 57.96 & 55.01 & 66.31  \cr  
Range Loss\cite{zhang2017range}         & 78.88 & 82.00 & 66.67 & 60.67 & 72.30  \cr
Class-Balanced Loss\cite{cui2019class}  & 73.88 & 76.77 & 60.28 & 60.58 & 67.45  \cr
OURS                                    & 80.32 & 78.11 & 70.05 & 68.52 & 75.18  \cr
    \toprule  \toprule    
    \end{tabular} 
 \end{threeparttable}   
\vspace{-0.2cm} 
\end{table*}

As seen, the performance favours the number of training samples in each subclass, while the few-shot subclasses fail to provide sufficient information for effective learning during training and for accurate prediction in testing. A direct way to alleviate the issue of data deficiency is to balance the data volume of each subclass via re-sampling. The under-sampling method is not a good choice since a large amount of useful information is ignored. By contrast, the over-sampling method shows a noticeable improvement in most cases, which suggests that the increasing volume of the majority subclasses could help the identification of the data in minority subclasses, under the condition that the resampled data is balanced. The focal loss achieves better performance than the re-sampling method due to the attention of the head and tail data. Considering that the samples obtained by data enhancement are the approximations of the existing samples and may have no significant contribution on model learning, the class-balanced loss tries to explicitly model the effective number of samples, improving the recognition accuracy. The range loss obtains the best performance among all the competitors, which punishes the influence of the extreme minority subclasses while enlarging the distances between different subclasses. This effectively prevents the minority categories from being swamped by the majority categories. The performance of all these methods does not reach the perfect level, especially for the subclasses with very limited samples. It is, hence, noted that CDLT provides an opportunity for algorithmic improvement in the long-tailed regime.

\subsection{Investigation on Concept Drift}
Here, we investigate the difficulty brought by the concept drift issue in the proposed dataset. Specifically, we evaluate the above-mentioned methods in two cases. The first case (denoted by "Original") states that the models are optimized on the normal training set, while the second case (denoted by "Drift") is that the concept drift data is ignored during training such that the models have no information about the drifted data. 
Experimental results are shown in Table \ref{Concept_Drift}, where the top-1 accuracy is given for each model.
We observe that all the models perform better in the first case than in the second case, and the two cases express a performance gap about 5\% for most models. Concept drift exhibit the changes of color and shape during the growth of the instances, in which the appearance features have been altered. If such concept drift data is not involved in training, the features of the instances extracted by the well-fitted models are hardly transferred to match the features in the drifted states. This verifies that the CDLT provides valuable data occurring concept drift for future practical research, and also poses the necessity of modelling the feature transferability of the concept drift in model design.
\begin{table}[htbp]\normalsize    
  \centering  
  \begin{threeparttable}  
  \renewcommand\tabcolsep{12.0pt}  
   \caption{The performance of the SOTA models on CDLT-cd. 
}  
  \label{Concept_Drift}  
   \begin{tabular}{c | c | cc}  
    \toprule \toprule  
	\textbf{Method}  & \textbf{Backbone} & \textbf{Ori.} & \textbf{Drift} \\  
    \hline  

NTS-Net\cite{Yang2018Learning}  & ResNet-50 & 74.32 & 69.00 \cr  
MCL\cite{chang2020devil}        & ResNet-50 & 71.78 & 67.14 \cr
MOMN\cite{min2020multi}         & ResNet-50 & 60.72 & 56.64 \cr
PMG~\cite{du2021progressive}    & ResNet-50 & 70.20 & 68.43 \cr
SPS~\cite{huang2021stochastic}  & ResNet-50 & 71.17 & 68.53 \cr
IMS~\cite{YE2023103837}         & ResNet-50 & 72.71 & 69.63 \cr 
OURS                            & ResNet-50 &\bf75.20 &\bf70.03 \cr
    \hline
ViT~\cite{dosovitskiy2020image}   & ViT-B-16 & 76.02  & 68.47 \cr
TransFG\cite{he2022transfg}       & ViT-B-16 & 76.63 & 69.51 \cr
FFVT~\cite{wang2021feature}       & ViT-B-16 & 75.61 & 69.58 \cr
IRM~\cite{arjovsky2019invariant}  & ViT-B-16 & 77.04 & 69.27 \cr
IB-IRM~\cite{ahuja2021invariance} & ViT-B-16 & 77.22 & 69.41 \cr
IMS~\cite{YE2023103837}           & ViT-B-16 & 77.92 & 70.82 \cr 
OURS                              & ViT-B-16 &\bf78.91 &\bf 71.06 \cr
    \toprule   
    \toprule  
    \end{tabular}  
    \end{threeparttable}  
\vspace{-0.1cm}  
\end{table} 
\vspace{-0.1cm}

\subsection{VLMs in Long-Tailed Distributions}\label{Ex_on_VLMs}
Vision-language models (VLMs) have demonstrated impressive performance in many downstream tasks. For example, CLIP achieved remarkable accuracy (60-80\%) on ImageNet. 
However, its performance in FGVC tasks warrants further investigation. We conducted experiments across six different datasets: CUB 200-2011, Stanford Dogs, and Oxford Pets, where each category has approximately equal training data, and NABirds, Stanford Cars, and CDLT, which exhibit pronounced long-tail distributions.

We use the results obtained by loading only the pre-trained parameters without any additional training as the pre-training baseline, obtaining only the number of classes in each dataset as prior knowledge. This approach aims to evaluate the performance of the model’s pre-trained knowledge across different visual concepts. 
Fine-tuning of CLIP is implemented by adding an Adapter to the text encoder. During training, a similarity matrix is calculated using the visual features of the current batch and the text features of all classes, with irrelevant text, such as label codes (e.g., n365697), removed from the dataset labels.

\begin{figure*}[htbp]
	\centering
	\begin{tabular}{ccc}
		\subfigure[CUB-200-2011]{\includegraphics[width=0.3 \linewidth]{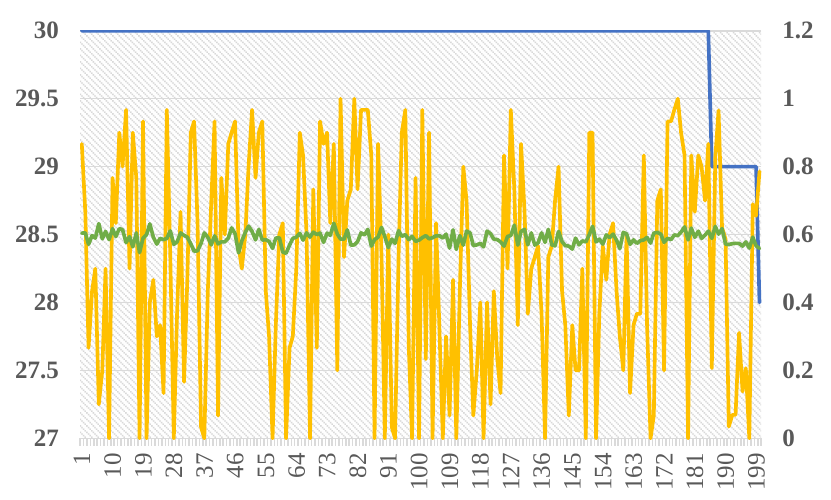}}\hspace{-0mm}
		&		
		\subfigure[Stanford Dogs]{\includegraphics[width=0.3 \linewidth]{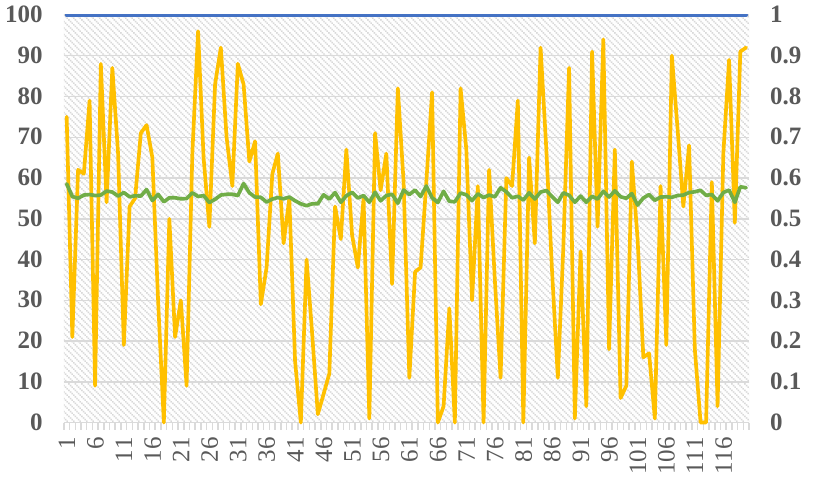}}\hspace{-0mm}
		&	
		\subfigure[Oxford Pet]{\includegraphics[width=0.3 \linewidth]{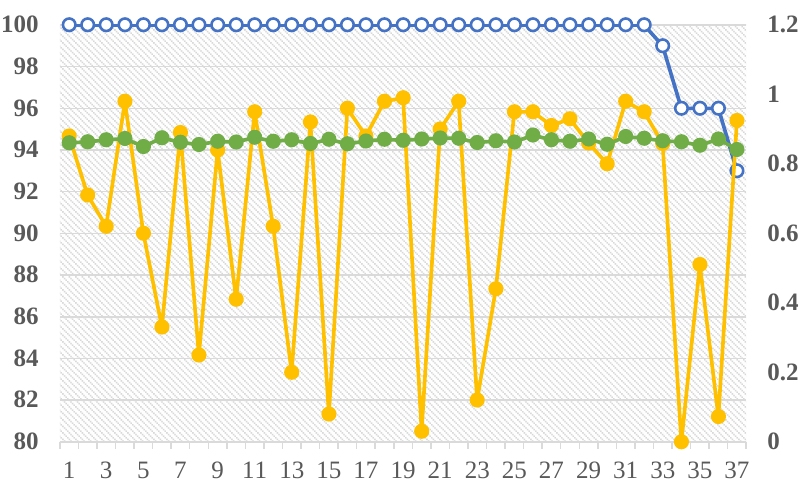}}\hspace{-0mm}
            \\   
            \subfigure[Aircraft]{\includegraphics[width=0.3 
        \linewidth]{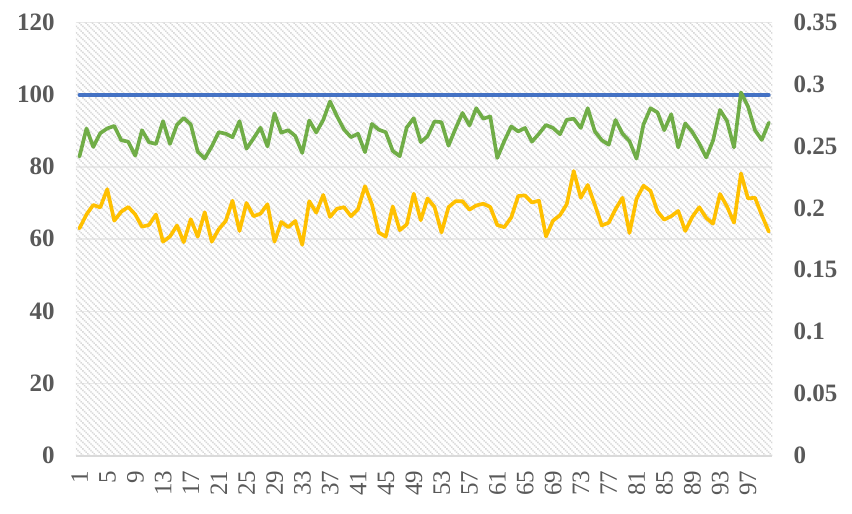}}\hspace{-0mm}
		&		
		\subfigure[CompCars]{\includegraphics[width=0.3 
        \linewidth]{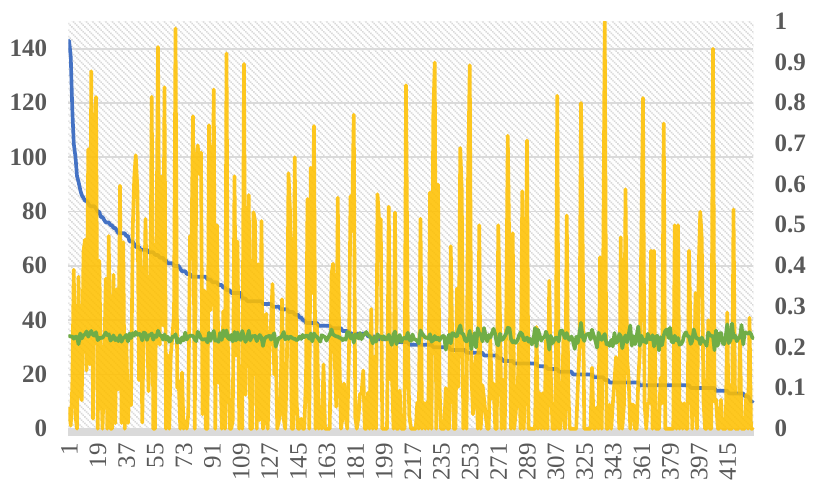}}\hspace{-0mm}
		&	
		\subfigure[Stanford Cars]{\includegraphics[width=0.3 \linewidth]{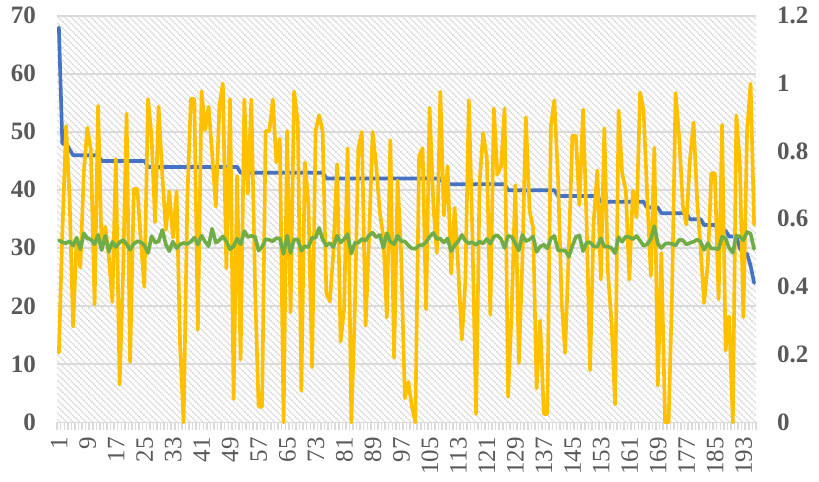}}\hspace{-0mm}
            \\ 
		\subfigure[NABirds]{\includegraphics[width=0.3 \linewidth]{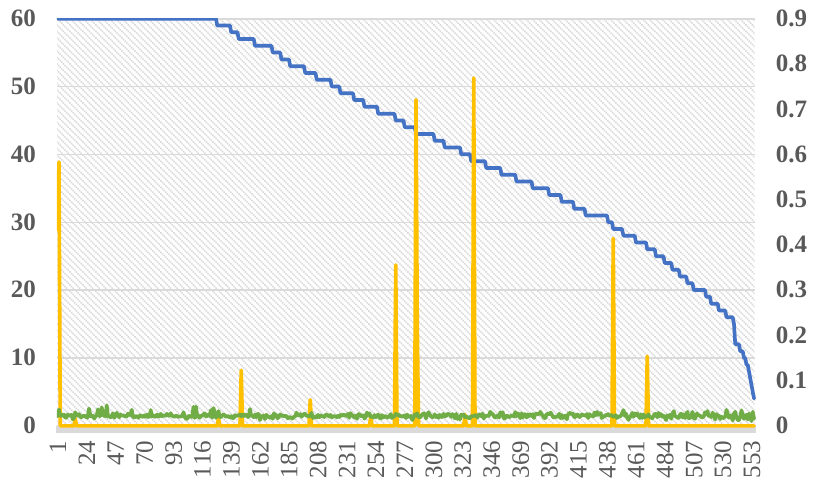}}\hspace{-0mm}
            &
		\subfigure[CDLT]{\includegraphics[width=0.3 \linewidth]{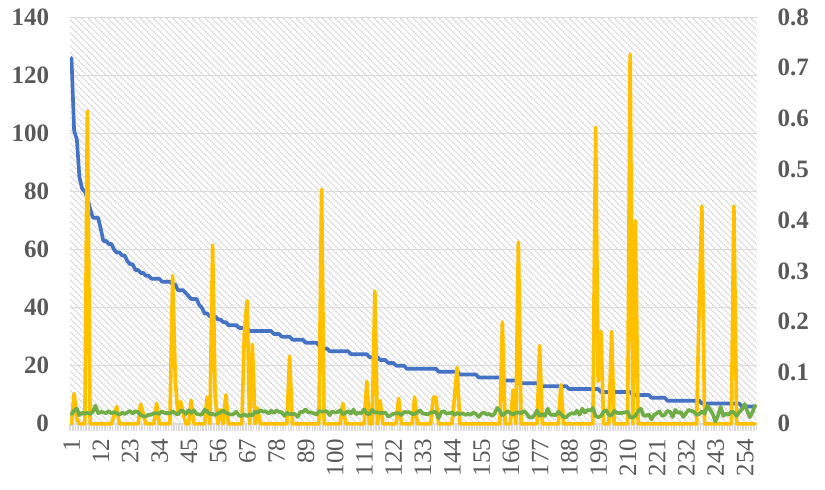}}\hspace{-0mm}\\
	\end{tabular}
\vspace{-0.2cm}
    \caption{Results of CLIP on different fine-grained datasets. The blue line represents the number of subclass images in the training set (values refer to the primary y-axis), while the yellow and green lines indicate CLIP's direct prediction results and fine-tuned prediction results for each subclass, respectively (accuracy values refer to the secondary y-axis).}
    \label{Visualization_of_CLIP}
\end{figure*}

Table~\ref{Pre_and_Finetuning} presents CLIP's results across different fine-grained datasets. It can be observed that the pre-trained knowledge is already sufficient to handle many common fine-grained targets. After fine-tuning, accuracy improves across four datasets. When using a larger pre-trained model, a significant improvement in accuracy was observed, demonstrating the impact of extensive pre-trained knowledge on downstream tasks.
However, the performance remains nearly unchanged on the NABirds and CDLT datasets.
We were curious about the reasons behind this performance difference, so we visualized the model's results for each subclass, as shown in the Figure~\ref{Visualization_of_CLIP}.

\begin{table}[htbp]\normalsize    
  \centering  
  \begin{threeparttable}  
  \renewcommand\tabcolsep{12.0pt}  
   \caption{Results of CLIP on different fine-grained datasets. "Pre." denotes using ViT-B as the backbone network without any additional training. "CLIP(B)" and "CLIP(L)" refer to models with ViT-B and ViT-L as backbone networks, respectively, trained using a fine-tuning strategy.}  
  \label{Pre_and_Finetuning}  
   \begin{tabular}{c|ccc}  
    \toprule \toprule  
	\textbf{Dataset}  & \textbf{Pre.} & \textbf{CLIP(B)} & \textbf{CLIP(L)}\\  
    \hline  
CUB 200-2011  &  50.89 & 58.91     & 67.03\cr
Stanford Dogs &  46.85 & 55.55     & 71.37\cr
Oxford Pets   &  68.37 & 83.60     & 87.62\cr 
Aircraft      &  19.60 & 26.22     & 43.06\cr
NAbirds       &  0.75  & 2.25      & 5.33\cr
CompCars      &  21.87 & 22.50     & 30.50 \cr
Stanford Cars &  57.73 & 53.22     & 72.65\cr 
CDLT          &  3.03  & 2.02      & 2.32 \cr
    \toprule   
    \toprule  
    \end{tabular}  
    \end{threeparttable}  
\vspace{-0.1cm}  
\end{table} 
\vspace{-0.1cm}

The model performs remarkably on well-defined visual concepts, such as "African hunting dog" in the Stanford Dogs dataset, achieving a classification accuracy of 92\% during testing. However, it demonstrates catastrophic misinterpretation on more ambiguous subclasses, like "redbone," with a test classification accuracy of only 0.09\%.
One reason for this phenomenon is that fine-grained labels often involve highly detailed descriptions to differentiate between categories. These descriptions are relatively uncommon and require a certain level of expertise, frequently involving various abbreviations. Additionally, the minimal differences between labels may hinder the model’s ability to extract precise category information.

Further analysis confirms that the fine-tuning process adjusts the model parameters to optimize overall accuracy. The model focuses on improving performance for categories that performed poorly on the training data, thereby avoiding the 'memorization' of certain feature patterns in the pre-trained knowledge. This approach alleviates the overfitting issues encountered during the upstream learning process.
However, when the dataset exhibits a long-tailed distribution, CLIP's performance shows certain biases. Fine-tuning has not significantly improved the recognition of categories with fewer samples, as tail-end subclasses are not sufficiently learned, making it challenging for the model to establish connections between these category labels and their semantic features. The reallocated weights may gradually shift the model's focus towards head classes, resulting in a 'dilution' of overfitted knowledge, while newly formed associations fall short of supporting accurate category learning, ultimately leading to a decline in overall performance.

Our visualization reveals three key insights:
(1) The pre-trained knowledge in VLMs is imbalanced for fine-grained tasks, showing significant performance disparities across different visual concepts.
(2) Long-tailed distributions exacerbate the learning difficulty for VLMs, resulting in stronger performance on common concepts and weaker performance on rare ones.
(3) This explains why recent generation-based large models tend to produce peculiar errors when applied to FGVC tasks.

\subsection{Future Work}

The above experimental results have shown that the emergence of concept drift and long-tailed distribution causes great difficulties to FGVC. For concept drift, the challenge arises from the changes of instance characteristics. It is worth noting that we do not use the auxiliary information about the season of the instance in the experiments. Proper use of the auxiliary information can effectively promote the real-world deployments of the FGVC algorithms. One possible solution is the multi-view learning, which tries to find the common features of the instance in different seasons and also fuse the complementary features to produce a full-view description of the instance. Another possible solution is conditional learning or stream learning, which regards the season information as a necessary time condition for generating the corresponding appearances and tries to find the common factors of different instances relating to season. Targeting at the long-tailed distribution issue, the head subclasses contribute a large number of features, which prevents the features of the tail subclasses from being well learned. A method worth trying is zero-shot learning, which reduces the dependence on the number of samples by modelling the attribute-level features that have high transferability between different subclasses. Another worth trial is to employ the generative model to enrich the samples of the minority subclasses, where the difficulty is improving the generative ability of the model with limited data. To summarize, the proposed CDLT is a good choice of dataset on advancing the research of real-world FGVC applications.


\section{Conclusions}
In this work, we build up a novel "CDLT" dataset to reveal the concept drift issue and the long-tailed distribution issue that are widely encountered in real FGVC scenarios. Extensive experiments benchmarked with the commonly used and the SOTA deep models demonstrate that CDLT is difficult to witness a pleasing performance, since the subclasses have high intra-class variation and high inter-class similarity.
We further conducted an in-depth investigation into the performance of recent popular VLMs on the CDLT dataset. Experimental results reveal significant disparities in the pre-trained knowledge of VLMs across diverse visual concepts, emphasizing the effects of long-tailed distributions on these models. Our dataset thus serves as a reliable benchmark for advancing research on concept drift and long-tailed distributions.

Although our work advances a step to bridge the gap in recognition capacity between machines and humans, more effects need to be taken on the concept drift and long-tailed distribution issues in practical FGVC. According to the current results, it is still difficult to determine which type of drift is more valuable to be researched for the development of the computer vision society, which drives us to provide specific drift labels for the samples in our future work.

\ifCLASSOPTIONcompsoc
  \section*{Acknowledgments}
\else
\fi
Thanks to Hongyu Ke, Ming Ma, Yue Li, Hao Men, Manqin Wen, Haodong Zhang, Ziyi Zhang, Xun Zhu, Yue Wu, Wenjin Hou, Mingxin Liao, Jiaqi Jiang, Huimin Xu, Ziming Hong, Shuhuang Chen, Yuxuan Zheng, and Wenzhou Chen for their contributions in the data collection process. Thanks to Yu Wang for their assistance in the data organization process.

\ifCLASSOPTIONcaptionsoff
  \newpage
\fi

\bibliographystyle{IEEEtran} 
\bibliography{ref.bib}


\begin{IEEEbiography}
[{\includegraphics[width=1in,height=1.25in,clip,keepaspectratio]{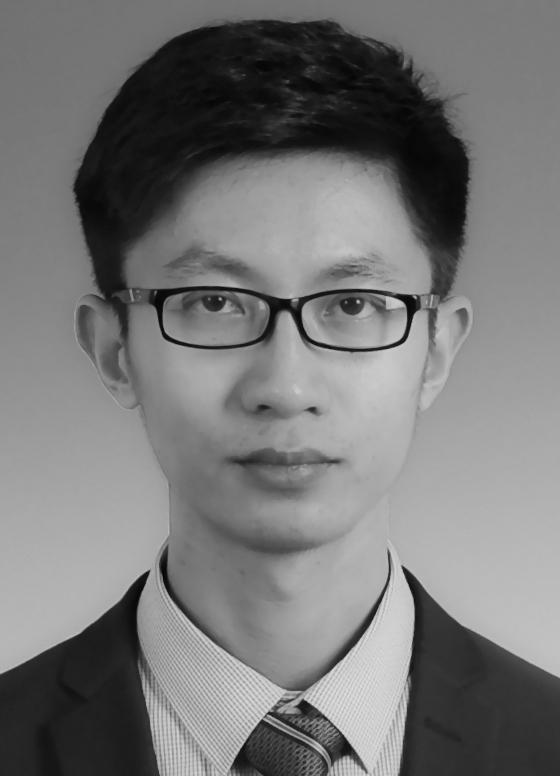}}]{Shuo Ye}
is currently a full-time Ph.D student in the School of Electronic Information and Communications, Huazhong University of Sciences and Technology (HUST), China. His current research interests span computer vision and voice signal processing with a series of topics, such as automatic speech recognition and fine-grained image categorization. 
\end{IEEEbiography}
\vspace{-1.1cm}

\begin{IEEEbiography}[{\includegraphics[width=1.15in,height=1.35in,clip,keepaspectratio]{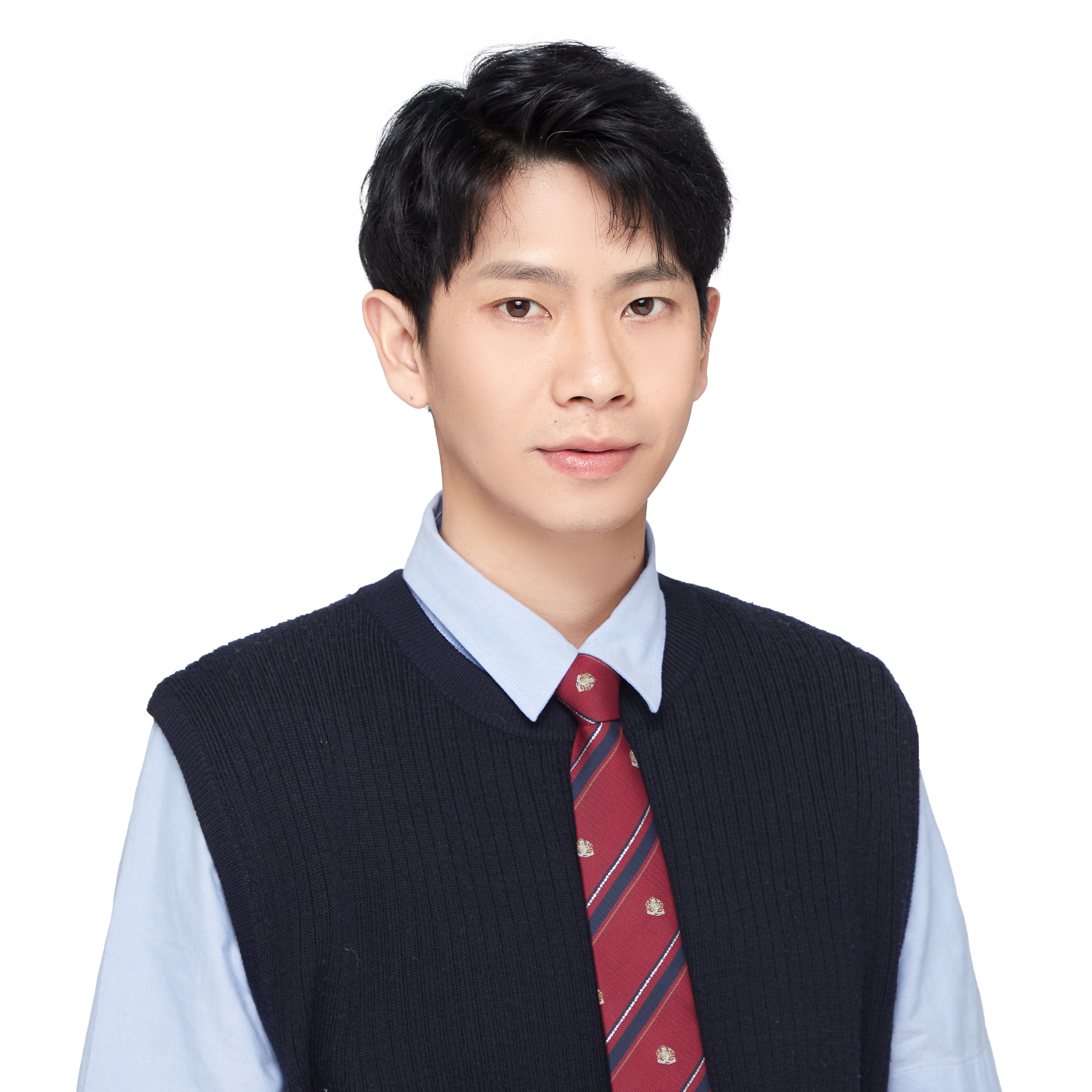}}]{Shiming Chen}
received the Ph.D. degree in the School of Electronic Information and Communications, Huazhong University of Sciences and Technology (HUST), China, in 2022. His research results have expounded in 20+ prominent conferences and prestigious journals, such as ICML, NeurIPS, ICCV, CVPR, IEEE TPAMI, IEEE TNNLS, and etc. His current research interests span computer vision and machine learning with a series of topics, such as generative modeling and learning, zero-shot learning, and visual-and-language learning.
\end{IEEEbiography}

\begin{IEEEbiography}[{\includegraphics[width=1in,height=1.25in,clip,keepaspectratio]{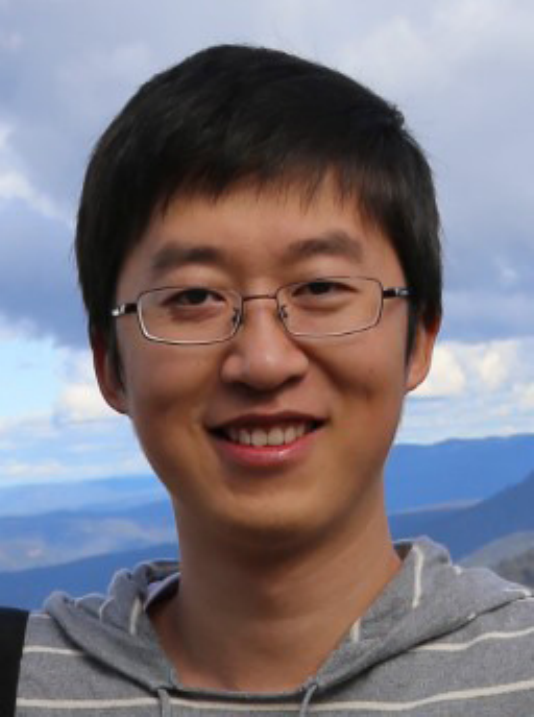}}]{Ruxin Wang}
is currently an associate professor with the School of Software, Yunnan University, Kunming, China. He received his BEng from Xidian University, his MSc from Huazhong University of Science and Technology, and his PhD degree from the University of Technology Sydney. He served as the chief technology officer in a startup. His research interests include image restoration, deep learning, and computer vision. He focuses on the topic of image synthesis by using both discriminative models and generative models. He has authored and co-authored 20+ research papers including IEEE T-NNLS, T-IP, T-Cyb, ICCV, and AAAI. He has received "the 1000 Talents Plan for Young Talents of Yunnan Province" award.
\end{IEEEbiography}

\begin{IEEEbiography}
[{\includegraphics[width=1in,height=1.25in,clip,keepaspectratio]{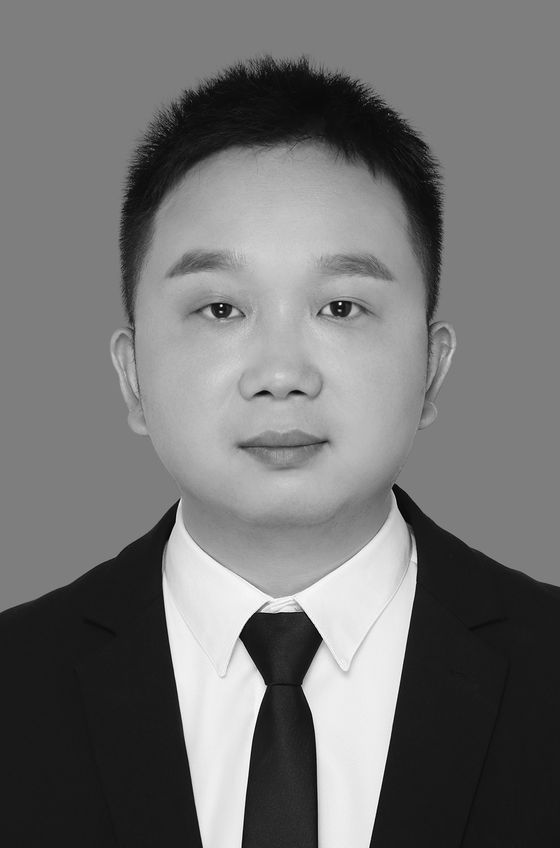}}]{Tianxu Wu}
is currently pursuing the Ph.D. in School of Electronic Information and Communications, Huazhong University of Science and Technology, Wuhan, China. His research interests include machine learning and computer vision.
\end{IEEEbiography}
\vspace{-1.1cm}

\begin{IEEEbiography}[{\includegraphics[width=1in,height=1.25in,clip,keepaspectratio]{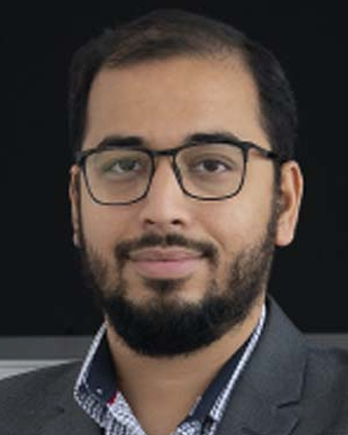}}]{Salman Khan}
(Senior Member, IEEE) received the PhD degree from The University of Western Aus tralia, in 2016. His PhD thesis received an honorable mentionon the Deans List Award. From 2016 to 2018, he was a research scientist at Data61 and CSIRO. He was a senior scientist with Inception Institute of Artificial Intelligence from 2018–2020. He is cur rently acting as an associate professor with Mohamed Bin Zayed University of Artificial Intelligence, since 2020, and an adjunct lecturer with Australian Na tional University, since 2016. He has served as a program committee member for several premier conferences, including CVPR, ICCV, IROS, ICRA, ICLR, NeurIPS, and ECCV. In 2019, he was awarded the outstanding reviewer award with CVPRandthebestpaperawardwithICPRAM 2020. He was the guest editor of IEEE Transactions on Pattern Analysis and Machine Intelligence. His research interests include computer vision and machine learning.
\end{IEEEbiography}

\begin{IEEEbiography}[{\includegraphics[width=1in,height=1.25in,clip,keepaspectratio]{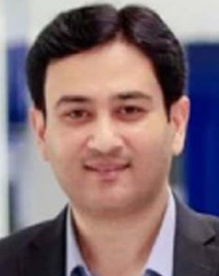}}]{Fahad Shahbaz Khan}
received the MSc degreein intelligent systems design from Chalmers University of Technology, Sweden, and the PhD degree in computer vision from the Autonomous University of Barcelona, Spain. He is a faculty member with MBZUAI, UAE and Linkoping University, Sweden. His research interests include a wide range of topics within computer vision, including object recognition, detection, segmentation, tracking and action recognition. He has achieved top ranks on various international challenges (Visual Object Tracking VOT: 1st 2014 and 2018, 2nd 2015, 1st 2016; VOT-TIR: 1st 2015 and 2016; OpenCV Tracking: 1st 2015; 1st PASCAL VOC Segmentation and Action Recognition tasks 2010). He received the best paper award in the computer vision track with IEEE ICPR 2016. He has published more than 100 reviewed conference papers, journal articles, and book contributions in these areas. He serves as a senior program committee member of several prestigious conferences, such as CVPR, AAAI, ACM Multimedia and associate editor of IEEE Transactions on Pattern Analysis and Machine Intelligence.
\end{IEEEbiography}

\begin{IEEEbiography}[{\includegraphics[width=1in,height=1.25in,clip,keepaspectratio]{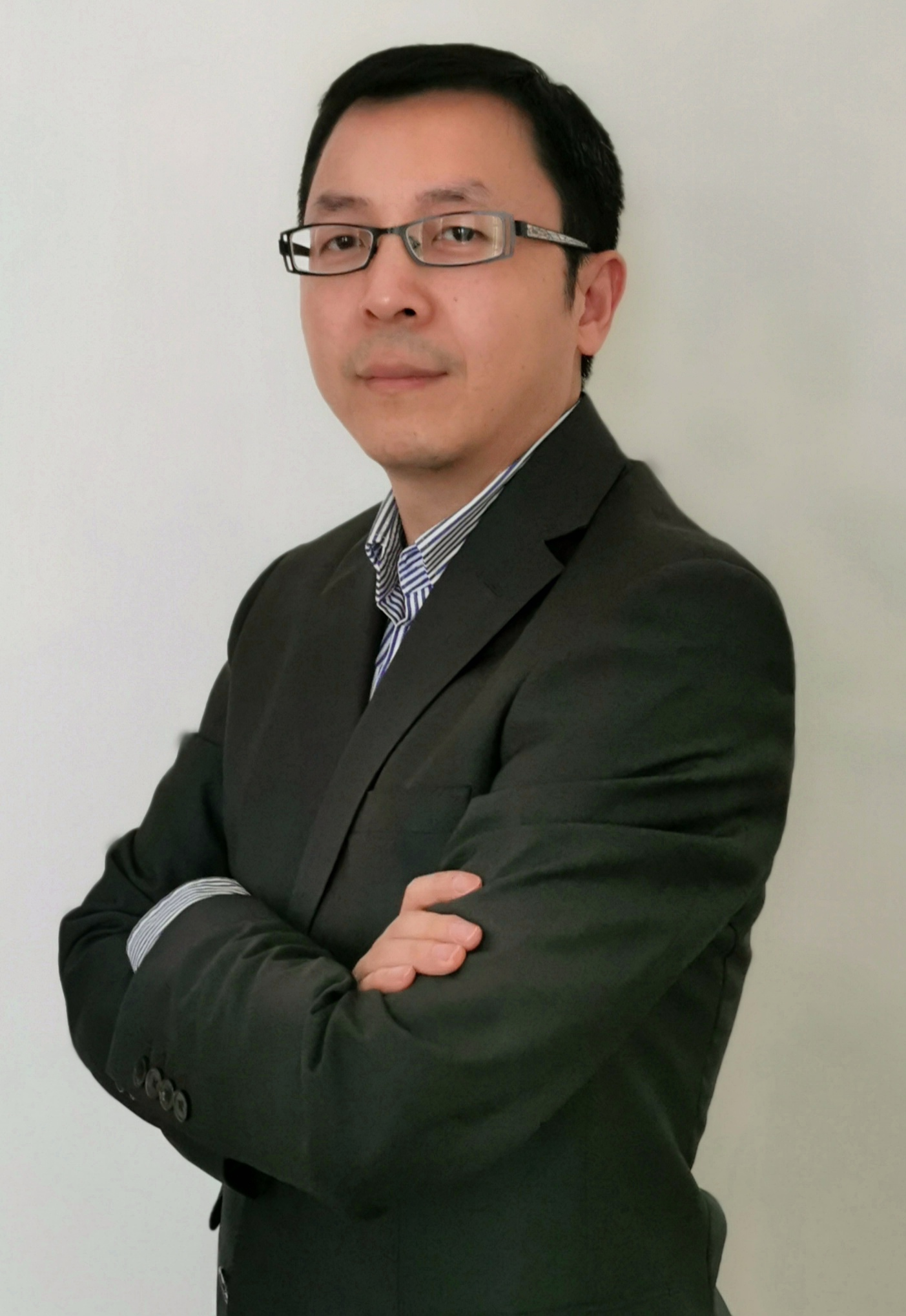}}]{Ling Shao}
(Fellow, IEEE)  is currently a Professor with the University of Chinese Academy of Sciences, Beijing, China. He received the Ph.D. degree in  Computer Vision, University of Oxford, England, in 2013. His research results have expounded in 400+ publications at prestigious journals and prominent conferences, such as IEEE T-PAMI, IJCV, T-IP, NeurIPS, ICCV, CVPR, ECCV. He has been an Associate Editor of \textit{IEEE Transactions on Image Processing}, \textit{IEEE Transactions on Neural Networks and Learning Systems}, \textit{IEEE Transactions on Circuits and Systems for Video Technology}, \textit{IEEE Transactions on Cybernetics}, and several other journals. He has edited four books and several special issues for journals such as IJCV. He has organized a number of international workshops with top conferences including ICCV, ECCV and ACM MM. He was the General Chair for BMVC 2018, has been an Area Chair or Program Committee member for many conferences, including ICCV, CVPR, ECCV and ACM MM.		
He was selected as a Highly Cited Researcher by the Web of Science in 2018-2024. He is a Fellow of the IEEE/IAPR/IET/British Computer Society. His research interests include computer vision and machine learning.
\end{IEEEbiography}

\end{document}